\documentclass[runningheads]{llncs}
\usepackage{graphicx}
\usepackage{tikz}
\usepackage{comment} 
\usepackage{amsmath,amssymb}
\usepackage{color}
\usepackage{algorithm}
\usepackage[noend]{algpseudocode}
\usepackage{url}
\usepackage{xspace}

\newcommand{\bx}{\boldsymbol{x}}

\newcommand{\bq}{\boldsymbol{q}}

\newcommand{\by}{\boldsymbol{y}}

\newcommand{\bg}{\boldsymbol{g}}
\newcommand{\bz}{\boldsymbol{z}}

\newcommand{\bv}{\boldsymbol{v}}

\newcommand{\sA}{\mathcal{A}}
\newcommand{\sC}{\mathcal{C}}
\newcommand{\sD}{\mathcal{D}}

\newcommand{\sL}{\mathcal{L}}
\newcommand{\bbO}{\mathbb{O}}

\newcommand{\etal}{\textit{et al}. }

\makeatletter
\DeclareRobustCommand\onedot{\futurelet\@let@token\@onedot}
\def\@onedot{\ifx\@let@token.\else.\null\fi\xspace}

\def\eg{\emph{e.g}\onedot} 
\def\ie{\emph{i.e}\onedot}

\def\etal{\emph{et al}\onedot}

\usepackage{xcolor}
\usepackage{multirow}

\makeatother
\makeatletter
\DeclareRobustCommand\onedot{\futurelet\@let@token\@onedot}
\def\@onedot{\ifx\@let@token.\else.\null\fi\xspace}

\def\eg{\emph{e.g}\onedot} 
\def\ie{\emph{i.e}\onedot}

\def\etal{\emph{et al}\onedot}
\makeatother

\begin{document}
\pagestyle{headings}
\mainmatter

\title{On the Effectiveness of Image Rotation\\for Open Set Domain Adaptation}

\titlerunning{On the Effectiveness of Image Rotation for Open Set Domain Adaptation}

\author{Silvia Bucci\thanks{equal contributions}\inst{,1,2}\orcidID{0000-0001-6318-7288} \and
Mohammad Reza Loghmani$^{\star}$\inst{,3}\orcidID{0000-0002-2687-7877} \and
Tatiana Tommasi\inst{1,2}\orcidID{0000-0001-8229-7159}}

\authorrunning{S. Bucci et al.}
\institute{Italian Institute of Technology, Italy\\ \and
Politecnico di Torino, Italy \\
\email{\{silvia.bucci,tatiana.tommasi\}@polito.it}\\ \and
Vision for Robotics laboratory, ACIN, TU Wien, 1040 Vienna, Austria\\
\email{loghmani@acin.tuwien.ac.at}}
\maketitle

\begin{abstract}
Open Set Domain Adaptation (OSDA) bridges the domain gap between a labeled source domain and an unlabeled target domain, while also rejecting target classes that are not present in the source.
To avoid negative transfer, OSDA can be tackled by first separating the known/unknown target samples and then aligning known target samples with the source data. We propose a novel method to addresses both these problems using the self-supervised task of rotation recognition. Moreover, we assess the performance with a new open set metric that properly balances the contribution of recognizing the known classes and rejecting the unknown samples.
Comparative experiments with existing OSDA methods on the standard Office-31 and Office-Home benchmarks show that: 
(i) our method outperforms its competitors, 
(ii) reproducibility for this field is a crucial issue to tackle, 
(iii) our metric provides a reliable tool to allow fair open set evaluation.
\end{abstract}

\section{Introduction}
\label{sec:intro}
The current success of deep learning models is showing how modern artificial intelligent systems can manage supervised machine learning tasks with growing accuracy. However, when the level of supervision decreases, all the limitations of the existing data-hungry approaches become evident. For many applications, large amount of supervised data are not readily available, moreover collecting and manually annotating such data may be difficult or very costly. Different sub-fields of computer vision, such as \emph{domain adaptation} \cite{csurka} and \emph{self-supervised learning} \cite{doersch2015unsupervised}, aim at designing new learning solutions to compensate for this lack of supervision. Domain adaptation focuses on leveraging a fully supervised data-rich source domain to learn a classification model that performs well on a different but related unlabeled target domain. 
Traditional domain adaptation methods assume that the target contains exactly the same set of labels of the source (\emph{closed-set} scenario). In recent years, this constraint has been relaxed in favor of the more realistic \emph{open-set} scenario where the target also contains samples drawn from unknown classes. In this case, it becomes important to identify and isolate the unknown class samples before reducing the domain shift to avoid negative transfer. Self-supervised learning focuses on training models on pretext tasks, such as image colorization or rotation prediction, using unlabeled data to then transfer the acquired high-level knowledge to new tasks with scarce supervision.
Recent literature has highlighted how self-supervision can be used for domain adaptation: jointly solving a pretext self-supervised task together with the main supervised problem leads to learning robust cross-domain features and supports generalization \cite{xu2019self-supervised,Carlucci_2019_CVPR}. Other works have also shown that the output of self-supervised models can be used in anomaly detection to discriminate normal and anomalous data  \cite{golan2018deep,Bergman2020Classification-Based}. However, these works only tackle binary problems (normal and anomalous class) and deal with a single domain. 

\begin{figure*}[t!]
    \centering
    \includegraphics[width=\linewidth]{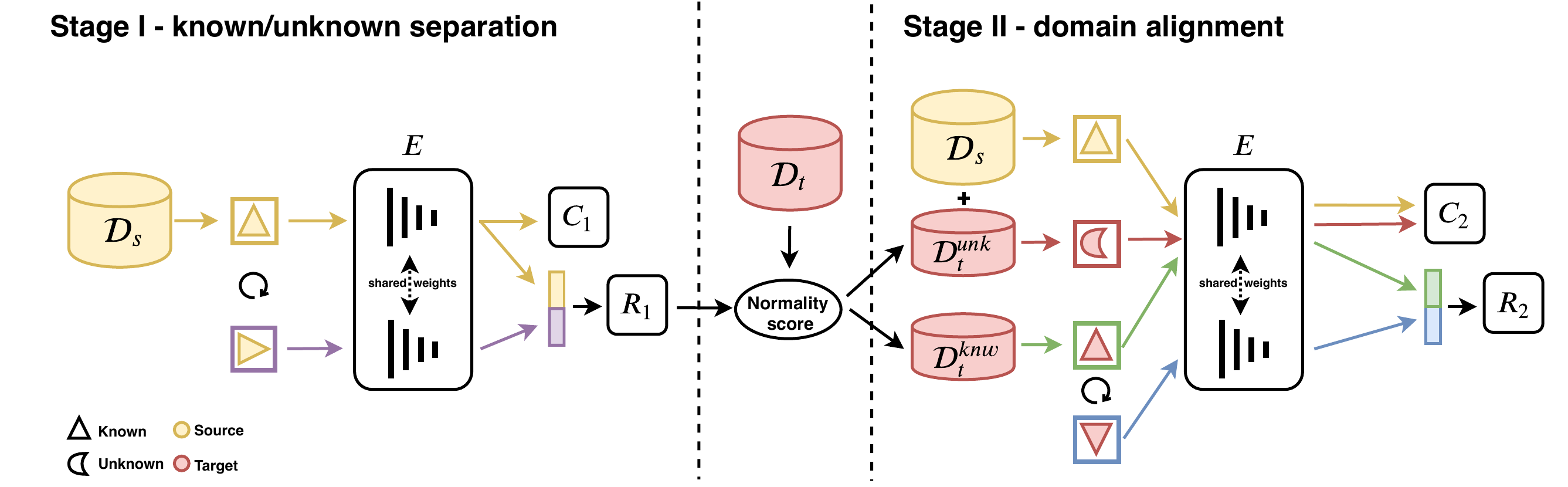}
    \caption{Schematic illustration of our Rotation-based Open Set (\textbf{ROS}). Stage I: the source dataset $\sD_s$ is used to train the encoder $E$, the semantic classifier $C_1$, and the multi-rotation classifier $R_1$ to perform known/unknown separation. $C_1$ is trained using the features of the original image, while $R_1$ is trained using the concatenated features of the original and rotated image. After convergence, the prediction of $R_1$ on the target dataset $\sD_t$ is used to generate a normality score that defines how the target samples are split into a known target dataset $\sD_t^{knw}$ and an unknown target dataset $\sD_t^{unk}$. Stage II: $E$, the semantic+unknown classifier $C_2$ and the rotation classifier $R_2$ are trained to align the source and target distributions and to recognize the known classes while rejecting the unknowns. $C_2$ is trained using the original images from $\sD_s$ and $\sD_t^{unk}$, while $R_2$ is trained using the concatenated features of the original and rotated known target samples.}
    \label{fig:scheme}
\end{figure*}

In this paper, we propose for the first time to use the inherent properties of self-supervision both for cross-domain robustness and for novelty detection to solve \emph{Open-Set Domain Adaptation} (OSDA). To this purpose, we propose a two-stage method called \emph{Rotation-based Open Set} (ROS) that is illustrated in Figure \ref{fig:scheme}. In the first stage, we separate the known and unknown target samples by training the model on a modified version of the rotation task that consists in predicting the relative rotation between a reference image and the rotated counterpart. In the second stage, we reduce the domain shift between the source domain and the known target domain using, once again, the rotation task. Finally we obtain a classifier that predicts each target sample as either belonging to one of the known classes or rejects it as unknown.
While evaluating ROS on the two popular benchmarks \emph{Office-31}~\cite{saenko2010adapting} and \emph{Office-Home}~\cite{venkateswara2017deep}, we expose the reproducibility problem of existing OSDA approaches and assess them with a new evaluation metric that better represents the performance of open set methods.
\noindent \textbf{We can summarize the contributions of our work as following}:
\begin{enumerate}
    \item we introduce a novel OSDA method that exploits rotation recognition to tackle both known/unknown target separation and domain alignment;
    \item we define a new OSDA metric that properly accounts for both known class recognition and unknown rejection;
    \item we present an extensive experimental benchmark against existing OSDA methods with two conclusions: (a) we put under the spotlight the urgent need of a rigorous experimental validation to guarantee result reproducibility; (b) our ROS defines the new state-of-the-art on two benchmark datasets.
\end{enumerate}
    
\noindent
A Pytorch implementation of our method, together with instructions to replicate our experiments, is available at \url{https://github.com/silvia1993/ROS} .

\section{Related Work}
\label{sec:related}
\textbf{Self-supervised learning} applies the techniques of supervised learning on problems where external supervision is not available. The idea is to manipulate the data to generate the supervision for an artificial task that is helpful to learn useful feature representations. Examples of self-supervised tasks in computer vision include predicting the relative position of image patches~\cite{doersch2015unsupervised,noroozi2016}, colorizing a gray-scale image~\cite{zhang2016colorful,Larsson_2017_CVPR}, and inpainting a removed patch~\cite{pathakCVPR16context}. Arguably, one of the most effective self-supervised tasks is rotation recognition~\cite{gidaris2018unsupervised} that consists in rotating the input images by multiples of $90^{\circ}$ and training the network to predict the rotation angle of each image. This pretext task has been successfully used in a variety of applications including anomaly detection~\cite{golan2018deep} and closed-set domain adaptation~\cite{xu2019self-supervised}. 

\textbf{Anomaly detection}, also known as outlier or novelty detection, aims at learning a model from a set of \emph{normal} samples to be able to detect out-of-distribution (\emph{anomalous}) instances. The research literature in this area is wide with three main kind of approaches. \emph{Distribution-based} methods~\cite{zimek2012survey,rKDE,deepanomaly,zong2018deep} model the distribution of the available normal data so that the anomalous samples can be recognized as those with a low likelihood under the learned probability function. \emph{Reconstruction-based} methods~\cite{Eskin2002,rPCA,Xia_2015_ICCV,autoenc,schlegl2017unsupervised} learn to reconstruct the normal samples from an embedding or a set of basis functions. Anomalous data are then recognized by having a larger reconstruction error with respect to normal samples. \emph{Discriminative} methods~\cite{OSVM,ruff18a,HendrycksG17,liang2018enhancing} train a classifier on the normal data and use its predictions to distinguish between normal and anomalous samples.

\textbf{Closed-set domain adaptation} (CSDA) accounts for the difference between source and target data by considering them as drawn from two different marginal distributions. The literature of DA can be divided into three groups based on the strategy used to reduce the domain shift. \emph{Discrepancy-based} methods~\cite{Long:2015,sun2016return,Xu_2019_ICCV} define a metric to measure the distance between source and target data in feature space. This metric is minimized while training the network to reduce the domain shift. \emph{Adversarial} methods~\cite{ganin2016domain,tzeng2017adversarial,russo2018from} aim at training a domain discriminator and a generator network in an adversarial fashion so that the generator converges to a solution that makes the source and target data indistinguishable for the domain discriminator. \emph{Self-supervised} methods~\cite{ghifary2016deep,bousmalis2016domain,Carlucci_2019_CVPR} train a network to solve an auxiliary self-supervised task on the target (and source) data, in addition to the main task, to learn robust cross-domain representations.

\textbf{Open Set Domain Adaptation} (OSDA) is a more realistic version of CSDA, where the source and target distribution do not contain the same categories. The term ``OSDA" was first introduced by Busto and Gall~\cite{Busto_2017_ICCV} that considered the setting where each domain contains, in addition to the shared categories, a set of private categories. The currently accepted definition of OSDA was introduced by Saito~\etal~\cite{saito2018open} that considered the target as containing all the source categories and additional set of private categories that should be considered \emph{unknown}. To date, only a handful of papers tackled this problem. \emph{Open Set Back-Propagation}~(OSBP)~\cite{saito2018open} is an adversarial method that consists in training a classifier to obtain a large boundary between source and target samples whereas the feature generator is trained to make the target samples far from the boundary. \emph{Separate To Adapt}~(STA)~\cite{liu2019separate} is an approach based on two stages. First, a multi-binary classifier trained on the source is used to estimate the similarity of target samples to the source. Then, target data with extreme high and low similarity are re-used to separate known and unknown classes while the features across domains are aligned through adversarial adaptation. \emph{Attract or Distract}~(AoD)~\cite{feng2019attract} starts with a mild alignment with a procedure similar to~\cite{saito2018open} and refines the decision by using metric learning to reduce the intra-class distance in known classes and push the unknown class away from the known classes. \emph{Universal Adaptation Network}~(UAN)\footnote{UAN is originally proposed for the universal domain adaptation setting that is a superset of OSDA, so it can also be used in the context of this paper.}~\cite{you2019universal} uses a pair of domain discriminators to both generate a sample-level transferability weight and to promote
the adaptation in the automatically discovered common label set. Differently from all existing OSDA methods, \textbf{our approach abandons adversarial training in favor of self-supervision. Indeed, we show that rotation recognition can be used, with tailored adjustments, both to separate known and unknown target samples and to align the known source and target distributions}\footnote{See Appendix E for a discussion on the use of other self-supervised tasks.}.

\section{Method}
\label{sec:method}
\subsection{Problem formulation}
\label{subsec:formulation}
Let us denote with $\mathcal{D}_s=\{(\bx_j^s,y_j^s)\}_{j=1}^{N_s} \sim p_s$ the labeled source dataset drawn from distribution $p_s$ and $\mathcal{D}_t=\{\bx^t_j\}_{j=1}^{N_t} \sim p_t$ the unlabeled target dataset drawn from distribution $p_t$. In OSDA, the source domain is associated with a set of \emph{known} classes $y^s \in \{1,\ldots, |\sC_s |\}$ that are shared with the target domain $\sC_s\subset\sC_t$, but the target covers also a set $\sC_{t \setminus s}$ of additional classes, which are considered \emph{unknown}. As in CSDA, it holds that $p_s\neq p_t$ and we further have that $p_s\neq p_t^{\sC_s}$, where $p_t^{\sC_s}$ denotes the distribution of the target domain belonging to the shared label space $\sC_s$. Therefore, in OSDA we face both a domain gap ($p_s\neq p_t^{\sC_s}$) and a category gap ($\sC_s \neq \sC_t$). OSDA approaches aim at assigning the target samples to either one of the $|\sC_s |$ shared classes or to reject them as \emph{unknown} using only annotated source samples, with the unlabeled target samples available transductively. An important measure characterizing a given OSDA problem is the \emph{openness} that relates the size of the source and target class set. For a dataset pair $(\sD_s, \sD_t )$, following the definition of~\cite{bendale2016towards}, the openness $\bbO$ is measured as $\bbO=1-\frac{|\sC_s |}{|\sC_t |}$. In CSDA $\bbO = 0$, while in OSDA $\bbO > 0$.

\subsection{Overview}
\label{subsec:overview}
When designing a method for OSDA, we face two main challenges: \emph{negative transfer} and \emph{known/unknown separation}. Negative transfer occurs when the whole source and target distribution are forcefully matched, thus also the unknown target samples are mistakenly aligned with source data. To avoid this issue, cross-domain adaptation should focus only on the shared $\sC_s$ classes, closing the gap between $p^{\sC_s}_t$ and $p_s$.
This leads to the challenge of known/unknown separation: recognizing each target sample as either belonging to one of the shared classes $\sC_s$ (known) or to one of the target private classes $\sC_{t \setminus s}$ (unknown). Following these observations, we structure our approach in two stages: (i) we separate the target samples into known and unknown, and (ii) we align the target samples predicted as known with the source samples (see Figure \ref{fig:scheme}). The first stage is formulated as an anomaly detection problem where the unknown samples are considered as anomalies. The second stage is formulated as a CSDA problem between source and the known target distribution. Inspired by recent advances in anomaly detection and CSDA~\cite{xu2019self-supervised,golan2018deep}, we solve both stages using the power of self-supervision. More specifically, we use two variations of the rotation classification task to compute a normality score for the known/unknown separation of the target samples and to reduce the domain gap.

\subsection{Rotation recognition for open set domain adaptation}
Let us denote with $rot90(\bx,i)$ the function that rotates clockwise a 2D image $\bx$ by $i\times90^{\circ}$. Rotation recognition is a self-supervised task that consists in rotating a given image $x$ by a random $i \in [1,4]$ and using a CNN to predict $i$ from the rotated image $\tilde{\bx}=rot90(\bx,i)$. 
We indicate with $|r|=4$ the cardinality of the label space for this classification task. 
In order to effectively apply rotation recognition to OSDA, we introduce the following variations.

\begin{figure}[t!]
\begin{minipage}[b]{0.42\textwidth}
\centering
\includegraphics[width=\textwidth]{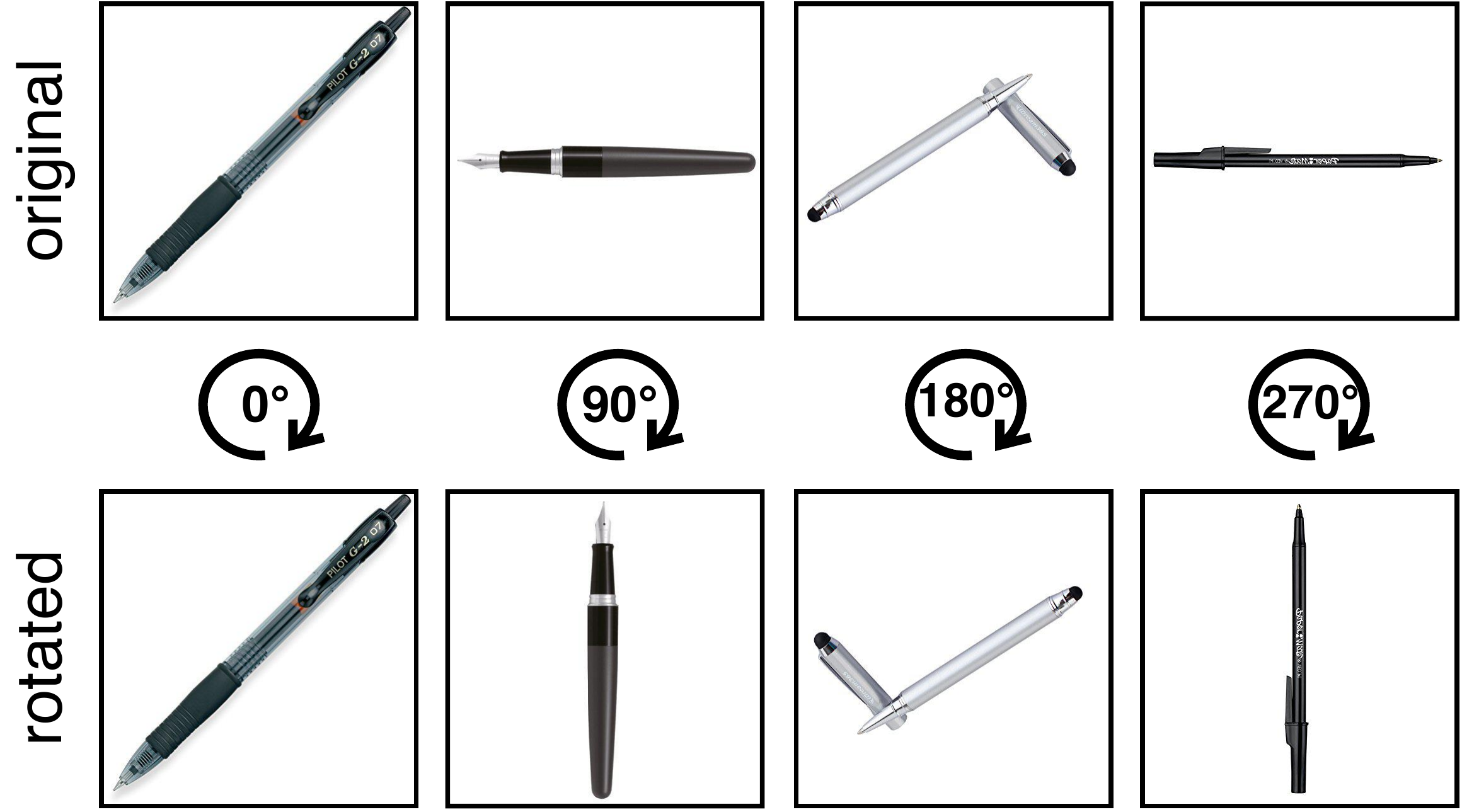}
\caption{\label{fig:relative_rotation} Are you able to infer the rotation degree of the rotated images without looking at the respective original one?}
\end{minipage}
\hfill
\begin{minipage}[b]{0.52\textwidth}
\centering
\includegraphics[width=\textwidth]{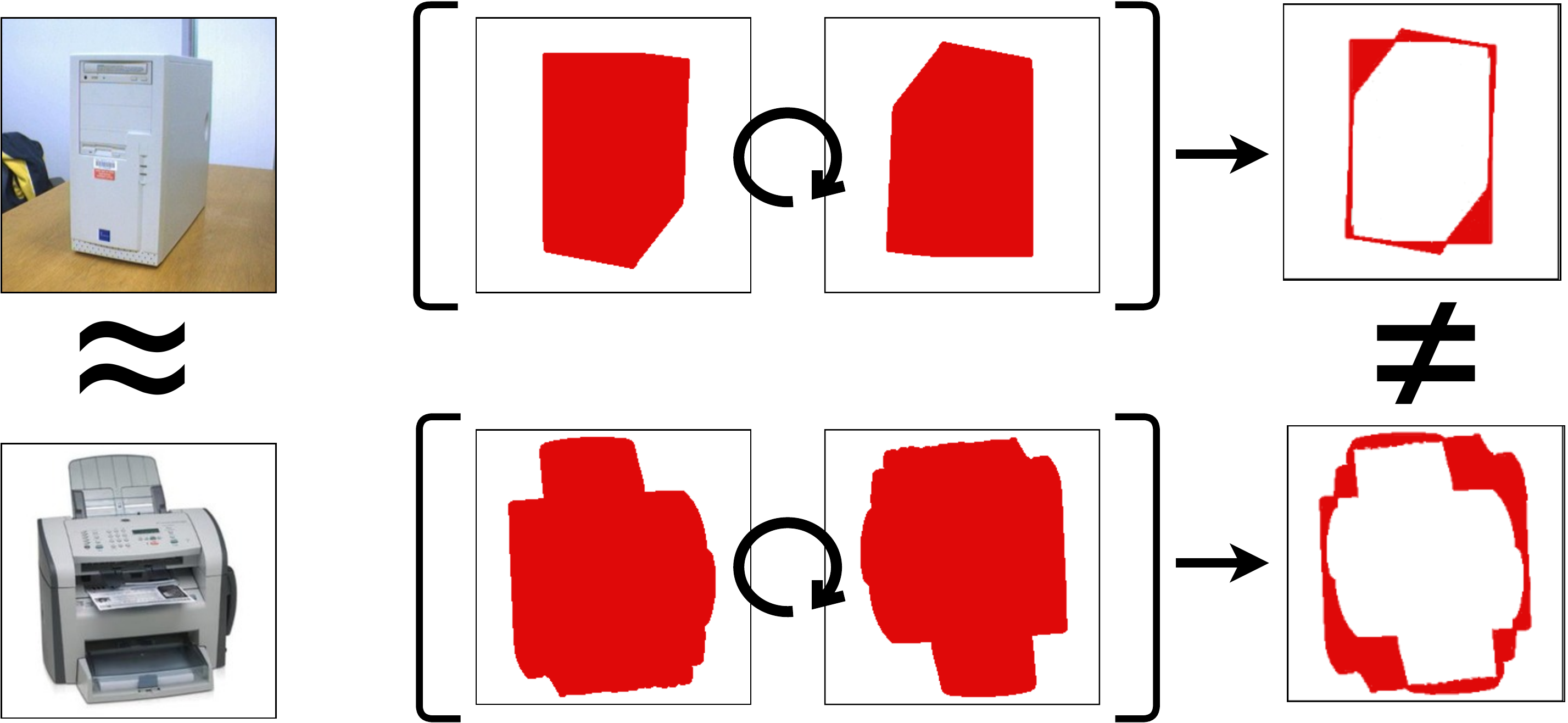}
\caption{\label{fig:relative_rotation2} The objects on the left may be confused. The relative rotation guides the network to focus on discriminative shape information}
\end{minipage}\vspace{-2mm}
\end{figure}

\textit{Relative rotation:}
Consider the images in Figure~\ref{fig:relative_rotation}. Inferring by how much each image has been rotated without looking at its original (non-rotated) version is an ill-posed problem since the pens, as all the other object classes, are not presented with a coherent orientation in the dataset. On the other hand, looking at both original and rotated image to infer the relative rotation between them is well-defined. Following this logic, we modify the standard rotation classification task~\cite{gidaris2018unsupervised} by introducing the original image as an anchor. Finally, we train the rotation classifier to predict the rotation angle given the concatenated features of both original (anchor) and rotated image. As indicated by Figure  \ref{fig:relative_rotation2}, the proposed relative rotation has the further effect of boosting the discriminative power of the learned features. It guides the network to focus more on specific shape details rather than on confusing texture information across different object classes.

\textit{Multi-rotation classification:} The standard setting of anomaly detection considers samples from one semantic category as the normal class and samples from other semantic categories as anomalies. Rotation recognition has been successfully applied to this setting, but it suffers when including multiple semantic categories in the normal class~\cite{golan2018deep}. This is the case when coping with the known/unknown separation of OSDA, where we have all the $|\sC_s|$ semantic categories as known data.
To overcome this problem, we propose a simple solution: we extend rotation recognition from a $4$-class problem to a $(4\times |\sC_s|)$-class problem, where the set of classes represents the combination of semantic and rotation labels. For example, if we rotate an image of category $y^s=2$ by $i=3$, its label for the multi-rotation classification task is $z^s=(y^s\times 4)+i=11$. 
In Appendix E, we discuss the specific merits of the multi-rotation classification task with further experimental evidences.
In the following, we indicate with $\by,\bz$ the one-hot vectors respectively for the class and multi-rotation labels.

\subsection{Stage I: known/unknown separation}
\label{subsec:stage1}
To distinguish between the known and unknown samples of $\sD_t$, we train a CNN on the multi-rotation classification task using $\tilde{\sD}_s=\{(\bx^s_j, \tilde{\bx}^s_j, z^s_j)\}^{4\times N_s}_{j=1}$. The network is composed of an encoder $E$ and two heads: a multi-rotation classifier $R_1$ and a semantic label classifier $C_1$.
The rotation prediction is computed on the stacked features of the original and rotated image produced by the encoder $\hat{\bz}^s=\text{softmax}\big(R_1([E(\bx^s),E(\tilde{\bx}^s)])\big)$, while the semantic prediction is computed only from the original image features as $\hat{\by}^s=\text{softmax}\big(C_1(E(\bx^s)\big)$. The network is trained to minimize the objective function $\sL_1 = \sL_{C_1} + \sL_{R_1}$, where the semantic loss $\sL_{C_1}$ is defined as a cross-entropy and the multi-rotation loss $\sL_{R_1}$ combines cross-entropy and center loss \cite{centerWenZL016}. More precisely,\vspace{-1mm}
\begin{align}
\sL_{C_1} &= -\sum_{j \in \sD_s} \by^s_j \cdot \log(\hat{\by}^s_j),\\
    \sL_{R_1} &= \sum_{j \in \tilde{\sD}_s} -\lambda_{1,1} \bz^s_j \cdot \log(\hat{\bz}^s_j) + \lambda_{1,2} ||\bv^s_j - \gamma(\bz^s_j)||^2_2, \label{eq:centerloss}\vspace{-6mm}
\end{align}
where $||.||_2$ indicates the $l_2$-norm operator, $\bv_j$ indicates the output of the penultimate layer of $R_1$ and $\gamma(\bz_j)$ indicates the corresponding centroid of the class associated with $\bv_j$. By using the center loss we further encourage the network to minimize the intra-class variations while keeping far the features of different classes. This supports the following use of the rotation classifier output as a metric to detect unknown category samples. 

Once the training is complete, we use $E$ and $R_1$ to compute the \emph{normality score} $\mathcal{N} \in [0,1]$ for each target sample, with large $\mathcal{N}$ values indicating normal (known) samples and vice-versa.
We start from the network prediction on all the relative rotation variants of a target sample $\hat{\bz_i}^t=\text{softmax}\big(R_1([E(\bx^t),E(\tilde{\bx}_i^t)])\big)_i$ and their related entropy $H(\hat{\bz}_i^t)= \big(\hat{\bz}_i^t \cdot \log(\hat{\bz}_i^t)/\log|\sC_s|\big)_i$ with $i=1,\ldots,|r|$. We indicate with $[\hat{\bz}^t]_m$ the $m$-th component of the $\hat{\bz}^t$ vector.
The full expression of the normality score is: 
\begin{equation}
    \mathcal{N}(\bx^t) = \max \Bigg\{ \max_{k=1,\ldots,|\sC_s|}\bigg(\sum_{i=1}^{|r|}[\hat{\bz}_i^t]_{k\times|r|+i}\bigg) , \bigg(1-\frac{1}{|r|}\sum_{i=1}^{|r|}H(\hat{\bz}_i^t)\bigg)\Bigg\}~.
    \label{eq:normalityscore}
\end{equation}
In words, this formula is a function of the ability of the network to correctly predict the semantic class and orientation of a target sample (first term in the braces, \emph{Rotation Score}) as well as of its confidence evaluated on the basis of the prediction entropy (second term, \emph{Entropy Score}). We maximize over these two components with the aim of taking the most reliable metric in each case.
Finally, the normality score is used to separate the target dataset into a known target dataset $\sD_t^{knw}$ and an unknown target dataset $\sD_t^{unk}$. The distinction is made directly through the data statistics using the average of the normality score over the whole target $\bar{\mathcal{N}}=\frac{1}{N_t}\sum_{j=1}^{N_t}\mathcal{N}_j$,
without the need to introduce any further parameter:  
\begin{equation}
\begin{cases}
    \bx^t \in \sD_t^{knw} & \quad \text{if} \quad \mathcal{N}(\bx^t) > \bar{\mathcal{N}} \\
    \bx^t \in \sD_t^{unk} & \quad  \text{if} \quad \mathcal{N}(\bx^t) < \bar{\mathcal{N}}~.
\end{cases}
\end{equation}
It is worth mentioning that only $R_1$ is directly involved in computing the normality score, while $C_1$ is only trained for regularization purposes and as a warm up for the following stage. For a detailed pseudo-code on how to compute $\mathcal{N}$ and generate $\sD_t^{knw}$ and $\sD_t^{unk}$, please refer to Appendix G.

\subsection{Stage II: domain alignment}
\label{subsec:stage2}
Once the target unknown samples have been identified, the scenario gets closer to that of standard CSDA. On the one hand, we can use $\sD_t^{knw}$ to close the domain gap without the risk of negative transfer and, on the other hand, we can exploit $\sD_t^{unk}$ to extend the original semantic classifier, making it able to recognize the unknown category.
Similarly to Stage I, the network is composed of an encoder $E$ and two heads: a rotation classifier $R_2$ and a semantic label classifier $C_2$. The encoder is inherited from the previous stage. The heads also leverage on the previous training phase but have two key differences with respect to Stage I:
(1) $C_1$ has a $|\sC_s|$-dimensional output, while $C_2$ has a $(|\sC_s|+1)$-dimensional output because of the addition of the unknown class; (2) $R_1$ is a multi-rotation classifier with a $(4\times|\sC_s|)$-dimensional output, $R_2$ is a rotation classifier with a $4$-dimensional output.
The rotation prediction is computed as $\hat{\bq}=\text{softmax}\big(R_2([E(\bx),E(\tilde{\bx})])\big)$  while the semantic prediction is  $\hat{\bg}=\text{softmax}\big(C_2(E(\bx)\big)$.
The network is trained to minimize the objective function $\sL_2 = \sL_{C_{2}} + \sL_{R_2}$, where $\sL_{C_{2}}$ combines the supervised cross-entropy and the unsupervised entropy loss for the classification task, while $\sL_{R_2}$ is defined as a cross-entropy for the rotation task.
The unsupervised entropy loss is used to involve in the semantic classification process also the unlabeled target samples recognized as known. This loss enforces the decision boundary to pass through low-density areas.
More precisely,
\begin{align}
\sL_{C_{2}} &= -\sum_{j \in \{\sD_{s}\cup \sD_t^{unk}\}} \bg_j \cdot \log(\hat{\bg}_j)
-\lambda_{2,1}\sum_{j \in \sD_t^{knw}} \hat{\bg}_j \cdot \log(\hat{\bg}_j),\\
\sL_{R_2} &=  -\lambda_{2,2}\sum_{j \in {\sD}_t^{knw}} \bq_j \cdot \log(\hat{\bq}_j)~.
\end{align}
Once the training is complete, $R_2$ is discarded and the target labels are simply predicted as $c_j^t=C_2(E(\bx_j^t))$ for all $j=1, \ldots, N_t$.

\section{On reproducibility and open set metrics}
\label{sec:metrics}
OSDA is a young field of research first introduced in 2017. As it is gaining momentum, it is crucial to guarantee the \emph{reproducibility} of the proposed methods and have a valid \emph{metric} to properly evaluate them.

\emph{Reproducibility:} In recent years, the machine learning community has become painfully aware of a reproducibility crisis~\cite{henderson2018deep,dodge2019show,lucic2018gans}. Replicating the results of state-of-the-art deep learning models is seldom straightforward due to a combination of non-deterministic factors in standard benchmark environments and poor reports from the authors. Although the problem is far from being solved, several efforts have been made to promote reproducibility through checklists~\cite{checklist}, challenges~\cite{challenge} and by encouraging authors to submit their code. On our side, we contribute by re-running the state-of-the-art methods for OSDA and compare them with the results reported in the papers (see Section~\ref{sec:expers}). Our results are produced using the original public implementation together with the parameters reported in the paper and, in some cases, repeated communications with the authors. We believe that this practice, as opposed to simply copying the results reported in the papers, can be of great value to the community.

\emph{Open set metrics:} The usual metrics adopted to evaluate OSDA are the average class accuracy over the known classes \emph{OS$^*$}, and the accuracy of the unknown class \emph{UNK}. They are generally combined in \emph{OS}$=\frac{|\sC_s |}{|\sC_s | + 1} \times$\emph{OS$^*$}$+ \frac{1}{|\sC_s | + 1} \times$\emph{UNK} as a measure of the overall performance.
However, we argue (and we already demonstrated in \cite{LOGHMANI2020198}) that treating the unknown as an additional class does not provide an appropriate metric.
As an example, let us consider an algorithm that is not designed to deal with unknown classes (\emph{UNK}=$0.0\%$) but has perfect accuracy over $10$ known classes (\emph{OS$^*$}=$100.0\%$). Although this algorithm is not suitable for open set scenarios because it completely disregards false positives,
it presents a high score of \emph{OS}=$90.9\%$. With increasing number of known classes, this effect on OS becomes even more acute, making the role of \emph{UNK} negligible. For this reason, we propose a new metric defined as the harmonic mean of \emph{OS$^*$} and \emph{UNK}, \emph{HOS}~$= 2 \frac{\text{\emph{OS}$^*$} \times \text{\emph{UNK}}}{\text{\emph{OS$^*$}} + \text{\emph{UNK}}}$. Differently from \emph{OS}, \emph{HOS} provides a high score only if the algorithm performs well both on known and on unknown samples, independently of $|\sC_s |$. Using a harmonic mean instead of a simple average penalizes large gaps between \emph{OS$^*$} and \emph{UNK}.

\section{Experiments}
\label{sec:expers}
\subsection{Setup: Baselines, Datasets}
We validate ROS with a thorough experimental analysis on two widely used benchmark datasets, Office-31 and Office-Home.
\emph{Office-31}~\cite{saenko2010adapting} consists of three domains, Webcam (W), Amazon (A) and Dslr (D), each containing $31$ object categories. We follow the setting proposed in~\cite{saito2018open}, where the first $10$ classes in alphabetic order are considered known and the last $11$ classes are considered unknown.
\emph{Office-Home}~\cite{venkateswara2017deep} consists of four domains, Product (Pr), Art (Ar), Real World (Rw) and  Clipart (Cl), each containing $65$ object categories.
Unless otherwise specified, we follow the setting proposed in~\cite{liu2019separate}, where the first $25$ classes in alphabetic order are considered known classes and the remaining $40$ classes are considered unknown. Both the number of categories and the large domain gaps make this dataset much more challenging than Office-31.

We compare ROS against the state-of-the-art methods STA~\cite{liu2019separate}, OSBP~\cite{saito2018open}, UAN~\cite{you2019universal}, AoD~\cite{feng2019attract},  that we already described in Section~\ref{sec:related}. For each of them, we run experiments using the official code provided by the authors, with the exact parameters declared in the relative paper. The only exception was made for AoD for which the authors have not released the code at the time of writing, thus we report the values presented in their original work.
We also highlight that STA presents a practical issue related to the similarity score used to separate known and unknown categories. Its formulation is based on the \emph{max} operator according to the Equation (2) in \cite{liu2019separate},
but appears instead based on \emph{sum} in the implementation code. In our analysis we considered both the two variants (STA\textsubscript{sum}, STA\textsubscript{max}) for the sake of completeness. All the results presented in this section, both for ROS and for the baseline methods, are the average over three independent experimental runs. We do not cherry pick the best out of several trials, but only run the three experiments we report.

\subsection{Implementation Details}
By following standard practice, we evaluate the performances of ROS on Office-31 using two different backbones {ResNet-50} \cite{he2016deep} and {VGGNet} \cite{simonyan2014very}, both pre-trained on ImageNet \cite{deng2009imagenet},  and we focus on ResNet-50 for Office-Home. The hyper-parameters values are the same regardless of the backbone and the dataset used. In particular, in both Stage I and Stage II of ROS the batch size is set to $32$ with a learning rate of $0.0003$ which decreases during the training following an inverse decay scheduling. For all layers trained from scratch, we set the learning rate $10$ times higher than the pre-trained ones.  We use SGD, setting the weight decay as $0.0005$ and momentum as $0.9$. In both stages, the weight of the self-supervised task is set three times the one of the semantic classification task, thus $\lambda_{1,1}= \lambda_{2,2}=3$. In Stage I, the weight of the center loss is  $\lambda_{1,2}=0.1$ and in Stage II the weight of the entropy loss is  $\lambda_{2,1}=0.1$. The network trained in Stage I is used as starting point for Stage II.
To take into consideration the extra category, in Stage II we set the learning rate of the new unknown class to twice that of the known classes (already learned in Stage I).
More implementation details and a sensitivity analysis of the hyper-parameters are provided in Appendix A and D.

\subsection{Results}
\label{subsec:results}

\paragraph{How does our method compare to the state-of-the-art?} 
Table~\ref{tab:office31Resnet50} and \ref{tab:officehomeResnet50} show the average results over three runs on each of the domain shifts, respectively of Office-31 and Office-Home. 
To discuss the results, we focus on the HOS metric since it is a synthesis of OS* and UNK, as discussed in Section \ref{sec:metrics}. Overall, ROS 
outperforms the state-of-the-art on a total of $13$ out of $18$ domain shifts and presents the highest average performance on both Office-31 and Office-Home.
The HOS improvement gets up to $2.2\%$ compared to the second best method OSBP. Specifically, ROS has a large gain over STA, regardless of its specific max or sum implementation, while UAN is not a challenging competitor due to its low performance on the unknown class. We can compare against AoD only when using VGG for Office-31: we report the  original results in gray in Table \ref{tab:officehomeResnet50}, with the HOS value confirming our advantage.

\begin{figure}
\centering
\includegraphics[width=0.85\linewidth]{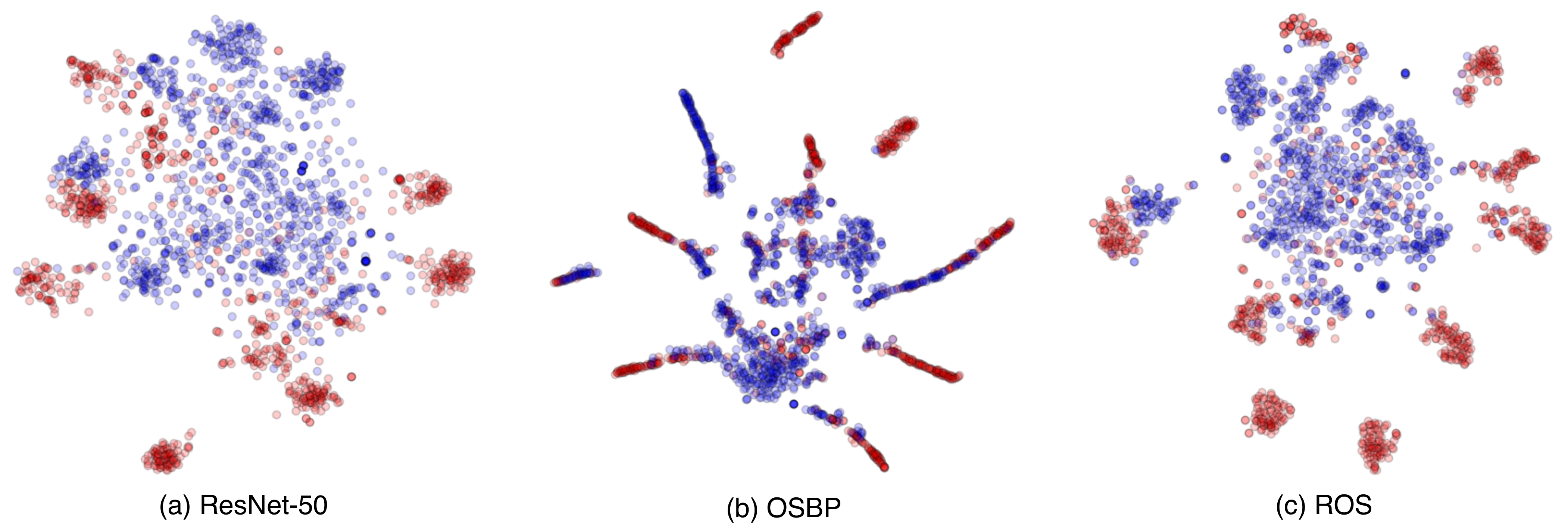}
\caption{t-SNE visualization of the target features for the
W$\rightarrow${A} domain shift from Office-31. Red and blue points are respectively features of known and unknown classes}
\label{fig:tsne}
\end{figure}

A more in-depth analysis indicates that the advantage of ROS is largely related in its ability in separating known and unknown samples.
Indeed, while our average OS* is similar to that of the competing methods, our average UNK is significantly higher. This characteristic is also visible qualitatively by looking at the t-SNE visualizations in Figure~\ref{fig:tsne} where we focus on the comparison against the second best method OSBP. Here the features for the known (red) and unknown (blue) target data appear more confused than for ROS.

\begin{table}[t!]
\centering
\caption{Accuracy (\%) averaged over three runs of each method on Office-31 dataset using ResNet-50 and VGGNet as backbones}
\resizebox{\textwidth}{!}{
\begin{tabular}{l@{~~~~~}cc ccc ccc ccc ccc ccc ccc ccc}
\hline

       &       \multicolumn{21}{c}{\textbf{Office-31}}        \\
       
\hline

&       \multicolumn{21}{c}{\textbf{ResNet-50}}        \\
\hline
 & &\multicolumn{3}{c|}{A $\rightarrow$ W } & \multicolumn{3}{c|}{A $\rightarrow$ D } & \multicolumn{3}{c|}{D $\rightarrow$ W }  & \multicolumn{3}{c|}{W $\rightarrow$ D} & \multicolumn{3}{c|}{D $\rightarrow$ A } & \multicolumn{3}{c|}{W $\rightarrow$ A } & \multicolumn{3}{c}{\textbf{Avg.} }
 \\
        & & OS* & UNK & \multicolumn{1}{c|}{\textbf{\underline{HOS}}} &  OS* & UNK &  \multicolumn{1}{c|}{\textbf{\underline{HOS}}}  & OS* & UNK &  \multicolumn{1}{c|}{\textbf{\underline{HOS}}}   & OS* & UNK &  \multicolumn{1}{c|}{\textbf{\underline{HOS}}}  & OS* & UNK &  \multicolumn{1}{c|}{\textbf{\underline{HOS}}}  & OS* & UNK &  \multicolumn{1}{c|}{\textbf{\underline{HOS}}}  & OS* & UNK &  \multicolumn{1}{c}{\textbf{\underline{HOS}}} \\
\hline
\multicolumn{1}{c}{STA\textsubscript{sum}}& \multirow{2}{*}{\cite{liu2019separate}}   & 92.1 & 58.0 & \multicolumn{1}{c|}{71.0}  & 95.4 & 45.5 & \multicolumn{1}{c|}{61.6}  & 97.1 & 49.7 & \multicolumn{1}{c|}{65.5}  & 96.6 & 48.5 & \multicolumn{1}{c|}{64.4} &  94.1 & 55.0 & \multicolumn{1}{c|}{69.4}  & 92.1 & 46.2 & \multicolumn{1}{c|}{60.9}  & 94.6 & 50.5 & \multicolumn{1}{c}{65.5$\pm$0.3}  \\

\multicolumn{1}{c}{STA\textsubscript{max}}&  & 86.7 & 67.6 & \multicolumn{1}{c|}{75.9}  & 91.0 & 63.9 & \multicolumn{1}{c|}{75.0}  & 94.1 & 55.5 & \multicolumn{1}{c|}{69.8}  & 84.9 & 67.8 & \multicolumn{1}{c|}{75.2}  & 83.1 & 65.9 & \multicolumn{1}{c|}{73.2} & 66.2 & 68.0 & \multicolumn{1}{c|}{66.1} & 84.3 & 64.8 & \multicolumn{1}{c}{72.5$\pm$0.8}\\

\multicolumn{1}{c}{OSBP}  \cite{saito2018open} & & 86.8 & 79.2 & \multicolumn{1}{c|}{\textbf{82.7}}  & 90.5 & 75.5 & \multicolumn{1}{c|}{\textbf{82.4}}  & 97.7 & 96.7 & \multicolumn{1}{c|}{\textbf{97.2}} & 99.1 & 84.2 & \multicolumn{1}{c|}{91.1} & 76.1 & 72.3 & \multicolumn{1}{c|}{75.1} &  73.0 & 74.4 & \multicolumn{1}{c|}{73.7}  & 87.2 & 80.4 & \multicolumn{1}{c}{83.7$\pm$0.4} \\

\multicolumn{1}{c}{UAN}   \cite{you2019universal}&   & 95.5 & 31.0 &  \multicolumn{1}{c|}{46.8}   & 95.6  & 24.4 & \multicolumn{1}{c|}{38.9}   & 99.8 & 52.5 & \multicolumn{1}{c|}{68.8}   & 81.5 & 41.4 & \multicolumn{1}{c|}{53.0}   & 93.5 & 53.4 & \multicolumn{1}{c|}{68.0}   & 94.1 & 38.8 & \multicolumn{1}{c|}{54.9}   & 93.4 & 40.3 & \multicolumn{1}{c}{55.1$\pm$1.4} \\

\multicolumn{1}{c}{\textbf{ROS}} &   & 88.4 & 76.7 & \multicolumn{1}{c|}{82.1}  & 87.5 & 77.8 & \multicolumn{1}{c|}{\textbf{82.4}}  & 99.3 & 93.0 & \multicolumn{1}{c|}{96.0}  & 100.0 & 99.4 & \multicolumn{1}{c|}{\textbf{99.7}}  & 74.8 & 81.2 & \multicolumn{1}{c|}{\textbf{77.9}}  & 69.7 & 86.6 & \multicolumn{1}{c|}{\textbf{77.2}}  & 86.6 & 85.8 & \multicolumn{1}{c}{\textbf{85.9$\pm$0.2}}   \\

\hline\hline

&       \multicolumn{21}{c}{\textbf{VGGNet}}        \\
\hline

&  &\multicolumn{3}{c|}{A $\rightarrow$ W } & \multicolumn{3}{c|}{A $\rightarrow$ D } & \multicolumn{3}{c|}{D $\rightarrow$ W }  & \multicolumn{3}{c|}{W $\rightarrow$ D} & \multicolumn{3}{c|}{D $\rightarrow$ A } & \multicolumn{3}{c|}{W $\rightarrow$ A } & \multicolumn{3}{c}{\textbf{Avg.} }
 \\
        & & OS* & UNK & \multicolumn{1}{c|}{\textbf{\underline{HOS}}} &  OS* & UNK &  \multicolumn{1}{c|}{\textbf{\underline{HOS}}}  & OS* & UNK &  \multicolumn{1}{c|}{\textbf{\underline{HOS}}}  &  OS* & UNK &  \multicolumn{1}{c|}{\textbf{\underline{HOS}}}  & OS* & UNK &  \multicolumn{1}{c|}{\textbf{\underline{HOS}}}  & OS* & UNK &  \multicolumn{1}{c|}{\textbf{\underline{HOS}}}  & OS* & UNK &  \multicolumn{1}{c}{\textbf{\underline{HOS}}} \\
\hline
\multicolumn{1}{c}{OSBP}  \cite{saito2018open}& & 79.4 & 75.8 & \multicolumn{1}{c|}{77.5}  & 87.9 & 75.2 & \multicolumn{1}{c|}{81.0}  & 96.8 & 93.4 & \multicolumn{1}{c|}{\textbf{95.0}}  & 98.9 & 84.2 & \multicolumn{1}{c|}{91.0} & 74.4 & 82.4 & \multicolumn{1}{c|}{\textbf{78.2}} & 69.7 & 76.4 & \multicolumn{1}{c|}{72.9}  & 84.5 & 81.2 & \multicolumn{1}{c}{82.6$\pm$0.8}   \\

\multicolumn{1}{c}{\textbf{ROS}} &  & 80.3 & 81.7 & \multicolumn{1}{c|}{\textbf{81.0}}  & 81.8 & 76.5 & \multicolumn{1}{c|}{79.0}  & 99.5 & 89.9 & \multicolumn{1}{c|}{94.4}  & 99.3 & 100.0 & \multicolumn{1}{c|}{\textbf{99.7}}  & 76.7  & 79.6 & \multicolumn{1}{c|}{78.1}  & 62.2 & 91.6 & \multicolumn{1}{c|}{\textbf{74.1}} & 83.3 & 86.5 & \multicolumn{1}{c}{\textbf{84.4$\pm$0.2}}    \\

\hline

\multicolumn{1}{c}{\textcolor{gray}{AoD \cite{feng2019attract}}}& & \textcolor{gray}{87.7} & \textcolor{gray}{73.4} & \multicolumn{1}{c|}{\textcolor{gray}{79.9}} &  \textcolor{gray}{92.0} & \textcolor{gray}{71.1} & \multicolumn{1}{c|}{\textbf{\textcolor{gray}{79.3}}}  & \textcolor{gray}{99.8} & \textcolor{gray}{78.9} & \multicolumn{1}{c|}{\textcolor{gray}{88.1}} & \textcolor{gray}{99.3} & \textcolor{gray}{87.2} & \multicolumn{1}{c|}{\textcolor{gray}{92.9}} &  \textcolor{gray}{88.4} & \textcolor{gray}{13.6} & \multicolumn{1}{c|}{\textcolor{gray}{23.6}}  & \textcolor{gray}{82.6} & \textcolor{gray}{57.3} & \multicolumn{1}{c|}{\textcolor{gray}{67.7}}   & \textcolor{gray}{91.6} & \textcolor{gray}{63.6} & \multicolumn{1}{c}{\textcolor{gray}{71.9}}  \\ 

\hline

\end{tabular} 
}
    \label{tab:office31Resnet50} 
\end{table}
\begin{table}[t]
\centering
\caption{Accuracy (\%) averaged over three runs of each method on Office-Home dataset using ResNet-50 as backbone}
\resizebox{\textwidth}{!}{
\begin{tabular}{l@{~~~~~~~}c ccc ccc ccc ccc ccc ccc ccc}
\hline

       &       \multicolumn{18}{c}{~~~~~~~~~~~\textbf{Office-Home}}        \\
       
\hline
 & & & \multicolumn{3}{c|}{Pr $\rightarrow$ Rw } & \multicolumn{3}{c|}{Pr $\rightarrow$ Cl } & \multicolumn{3}{c|}{Pr $\rightarrow$ Ar }  & 
 
 \multicolumn{3}{c|}{Ar $\rightarrow$ Pr } & \multicolumn{3}{c|}{Ar $\rightarrow$ Rw } & \multicolumn{3}{c}{Ar $\rightarrow$ Cl }  \\
       & &  & OS* & UNK & \multicolumn{1}{c|}{\textbf{\underline{HOS}}} &  OS* & UNK &  \multicolumn{1}{c|}{\textbf{\underline{HOS}}} &  OS* & UNK &  \multicolumn{1}{c|}{\textbf{\underline{HOS}}} &  OS* & UNK &  \multicolumn{1}{c|}{\textbf{\underline{HOS}}} &  OS* & UNK &  \multicolumn{1}{c|}{\textbf{\underline{HOS}}} &  OS* & UNK &  \multicolumn{1}{c}{\textbf{\underline{HOS}}} \\
\hline

\multicolumn{1}{c}{STA\textsubscript{sum}} & \multirow{2}{*}{\cite{liu2019separate}}  & &  78.1 & 63.3 & \multicolumn{1}{c|}{69.7}  & 44.7 & 71.5 & \multicolumn{1}{c|}{55.0} & 55.4 & 73.7 & \multicolumn{1}{c|}{63.1} & 68.7 & 59.7 & \multicolumn{1}{c|}{63.7}  & 81.1 & 50.5 & \multicolumn{1}{c|}{62.1}  & 50.8 & 63.4 & \multicolumn{1}{c}{56.3}  \\
\multicolumn{1}{c}{STA\textsubscript{max}}& &  & 76.2 & 64.3 & \multicolumn{1}{c|}{69.5}  & 44.2 & 67.1 & \multicolumn{1}{c|}{53.2}  & 54.2 & 72.4 & \multicolumn{1}{c|}{61.9}  & 68.0 & 48.4 & \multicolumn{1}{c|}{54.0}   & 78.6 & 60.4 & \multicolumn{1}{c|}{68.3}  & 46.0 & 72.3 & \multicolumn{1}{c}{55.8}\\
\multicolumn{1}{c}{OSBP} \cite{saito2018open}& &   & 76.2 & 71.7 & \multicolumn{1}{c|}{73.9}  & 44.5 & 66.3 & \multicolumn{1}{c|}{53.2}  & 59.1 & 68.1 & \multicolumn{1}{c|}{\textbf{63.2}}  & 71.8 & 59.8 & \multicolumn{1}{c|}{65.2}  & 79.3 & 67.5 & \multicolumn{1}{c|}{72.9}   & 50.2 & 61.1 & \multicolumn{1}{c}{55.1}\\
\multicolumn{1}{c}{UAN} \cite{you2019universal}&  &  & 84.0 & 0.1 & \multicolumn{1}{c|}{0.2}  & 59.1 & 0.0 & \multicolumn{1}{c|}{0.0}  & 73.7 & 0.0 & \multicolumn{1}{c|}{0.0}  & 81.1 & 0.0 & \multicolumn{1}{c|}{0.0}  & 88.2 & 0.1 & \multicolumn{1}{c|}{0.2}  & 62.4 & 0.0 & \multicolumn{1}{c}{0.0} \\
\multicolumn{1}{c}{\textbf{ROS}} & & & 70.8 & 78.4 &  \multicolumn{1}{c|}{\textbf{74.4}} & 46.5 & 71.2 & \multicolumn{1}{c|}{\textbf{56.3}}  & 57.3 & 64.3 & \multicolumn{1}{c|}{60.6}  & 68.4 & 70.3 & \multicolumn{1}{c|}{\textbf{69.3}} & 75.8 & 77.2 & \multicolumn{1}{c|}{\textbf{76.5}} & 50.6 & 74.1 & \multicolumn{1}{c}{\textbf{60.1}} \\
\hline\hline
 & \multicolumn{3}{c|}{Rw $\rightarrow$ Ar } & \multicolumn{3}{c|}{Rw $\rightarrow$ Pr } & \multicolumn{3}{c|}{Rw $\rightarrow$ Cl }
 
  & \multicolumn{3}{c|}{Cl $\rightarrow$ Rw } & \multicolumn{3}{c|}{Cl $\rightarrow$ Ar } & \multicolumn{3}{c}{Cl $\rightarrow$ Pr }  & \multicolumn{3}{c}{\textbf{Avg.}} \\ 
  
          & OS* & UNK & \multicolumn{1}{c|}{\textbf{\underline{HOS}}} &  OS* & UNK &  \multicolumn{1}{c|}{\textbf{\underline{HOS}}} &  OS* & UNK &  \multicolumn{1}{c|}{\textbf{\underline{HOS}}}&  OS* & UNK &  \multicolumn{1}{c|}{\textbf{\underline{HOS}}}  & OS* & UNK &  \multicolumn{1}{c|}{\textbf{\underline{HOS}}}  & OS* & UNK &  \multicolumn{1}{c|}{\textbf{\underline{HOS}}} & OS* & UNK &  \multicolumn{1}{c}{\textbf{\underline{HOS}}} \\
         
\hline
\multicolumn{1}{c}{STA\textsubscript{sum}}  & 67.9 & 62.3 & \multicolumn{1}{c|}{65.0}  & 77.9 & 58.0 & \multicolumn{1}{c|}{66.4}  & 51.4 & 57.9 & \multicolumn{1}{c|}{54.2}  & 69.8 & 63.2 & \multicolumn{1}{c|}{66.3}   & 53.0 & 63.9 & \multicolumn{1}{c|}{57.9}  & 61.4 & 63.5 & \multicolumn{1}{c|}{62.5}  & 63.4 & 62.6 & \multicolumn{1}{c}{61.9$\pm$2.1}  \\
\multicolumn{1}{c}{STA\textsubscript{max}} & 67.5 & 66.7 & \multicolumn{1}{c|}{67.1}  & 77.1 & 55.4 & \multicolumn{1}{c|}{64.5} & 49.9 & 61.1 & \multicolumn{1}{c|}{54.5} & 67.0 & 66.7 & \multicolumn{1}{c|}{66.8}   & 51.4 & 65.0 & \multicolumn{1}{c|}{57.4}   & 61.8 & 59.1 & \multicolumn{1}{c|}{60.4}  & 61.8 & 63.3 & \multicolumn{1}{c}{61.1$\pm$0.3} \\
\multicolumn{1}{c}{OSBP}  & 66.1 & 67.3 & \multicolumn{1}{c|}{66.7} &  76.3 & 68.6 & \multicolumn{1}{c|}{72.3}  & 48.0 & 63.0 & \multicolumn{1}{c|}{54.5}  & 72.0 & 69.2 & \multicolumn{1}{c|}{\textbf{70.6}}   & 59.4 & 70.3 & \multicolumn{1}{c|}{\textbf{64.3}}   & 67.0 & 62.7 & \multicolumn{1}{c|}{64.7}    & 64.1 & 66.3 & \multicolumn{1}{c}{64.7$\pm$0.2} \\
\multicolumn{1}{c}{UAN}   & 77.5 & 0.1 & \multicolumn{1}{c|}{0.2}  & 85.0 & 0.1 & \multicolumn{1}{c|}{0.1}  & 66.2 & 0.0 & \multicolumn{1}{c|}{0.0}  & 80.6 & 0.1 & \multicolumn{1}{c|}{0.2}  & 70.5 & 0.0 & \multicolumn{1}{c|}{0.0}  & 74.0 & 0.1 & \multicolumn{1}{c|}{0.2}   & 75.2 & 0.0 & \multicolumn{1}{c}{0.1$\pm$0.0} \\
\multicolumn{1}{c}{\textbf{ROS}}  & 67.0 & 70.8 & \multicolumn{1}{c|}{\textbf{68.8}} & 72.0 & 80.0 & \multicolumn{1}{c|}{\textbf{75.7}}  & 51.5 & 73.0 & \multicolumn{1}{c|}{\textbf{60.4}}  & 65.3 & 72.2 & \multicolumn{1}{c|}{68.6}  & 53.6 & 65.5 & \multicolumn{1}{c|}{58.9} & 59.8 & 71.6 & \multicolumn{1}{c|}{\textbf{65.2}} & 61.6 & 72.4 & \multicolumn{1}{c}{\textbf{66.2$\pm$ 0.3}}\\       
         
\hline

\end{tabular}
}
    \label{tab:officehomeResnet50} 
\end{table}

\paragraph{Is it possible to reproduce the reported results of the state-of-the-art?}
By analyzing the published OSDA papers, we noticed some incoherence in the reported results. For example, some of the results from OSBP are different between the pre-print~\cite{saito2018open-arxiv} and the published~\cite{saito2018open} version, although they present the same description for method and hyper-parameters. Also, AoD~\cite{feng2019attract} compares against the pre-print results of OSBP, while omitting the results of STA. To dissipate these ambiguities and gain a better perspective on the current state-of-the-art methods, in Table~\ref{tab:reproducibility} we compare the results on Office-31 reported in previous works with the results obtained by running their code. For this analysis we focus on OS since it is the only metric reported for some of the methods. The comparison shows that, despite using the original implementation and the information provided by the authors, the OS obtained by re-running the experiments is between $1.3\%$ and $4.9\%$ lower than the originally published results. The significance of this gap calls for greater attention in providing all the relevant information for reproducing the experimental results. A more extensive reproducibility study is provided in Appendix B.
\begin{table}[t]
\centering
\caption{Reported vs reproduced OS accuracy (\%) averaged over three runs}
\resizebox{0.8\textwidth}{!}{
\begin{tabular}{l@{~~~} cccccc ccc}
\hline
 \multicolumn{9}{c}{\textbf{Reproducibility Study}} \\
 \hline
 \multicolumn{6}{c|}{Office-31 (ResNet-50)} & \multicolumn{3}{c}{Office-31 (VGGNet)}\\
 \hline
  \multicolumn{3}{c|}{STA\textsubscript{sum}} & \multicolumn{3}{c|}{UAN}  & \multicolumn{3}{c}{OSBP}\\
OS\textsubscript{reported} & OS\textsubscript{ours} & \multicolumn{1}{c|}{gap} & OS\textsubscript{reported} & OS\textsubscript{ours}  & \multicolumn{1}{c|}{gap} & OS\textsubscript{reported} & OS\textsubscript{ours}  & \multicolumn{1}{c}{gap} \\
 \hline
 \multicolumn{1}{c}{92.9} &  \multicolumn{1}{c}{90.6$\pm$1.8} & \multicolumn{1}{c|}{\textbf{2.3}} &  \multicolumn{1}{c}{89.2} &   \multicolumn{1}{c}{87.9$\pm$0.03} & \multicolumn{1}{c|}{\textbf{1.3}} &  \multicolumn{1}{c}{89.1} &  \multicolumn{1}{c}{84.2 $\pm$0.4} & \multicolumn{1}{c}{\textbf{4.9}}\\
\hline
\end{tabular}
}
    \label{tab:reproducibility} 
\end{table}

\paragraph{Why is it important to use the HOS metric?}
The most glaring example of why OS is not an appropriate metric for OSDA is provided by the results of UAN. In fact, when computing OS from the average (OS*,UNK) in Table~\ref{tab:office31Resnet50} and~\ref{tab:officehomeResnet50}, we can see that UAN has OS=$72.5\%$ for Office-Home and OS=$91.4\%$ for Office-31. This is mostly reflective of the ability of UAN in recognizing the known classes (OS*), but it completely disregards its (in)ability to identify the unknown samples (UNK). For example, for most domain shifts in Office-Home, UAN does not assign (almost) any samples to the unknown class, resulting in UNK=$0.0\%$. On the other hand, HOS better reflects the open set scenario and assumes a high value only when OS* and UNK are both high.

\paragraph{Is rotation recognition effective for known/unknown separation in OSDA?}
To better understand the effectiveness of rotation recognition for known/unknown separation, we measure the performance of our Stage I and compare it to the Stage I of STA. Indeed, also STA has a similar two-stage structure, but uses a multi-binary classifier instead of a multi-rotation classifier to separate known and unknown target samples. To assess the performance, we compute the \textit{area under receiver operating characteristic curve} (AUC-ROC) over the normality scores $\mathcal{N}$ on Office-31. Table~\ref{tab:ablationstep1} shows that the AUC-ROC of ROS ($91.5$) is significantly higher than that of the multi-binary used by STA ($79.9$). Table~\ref{tab:ablationstep1} also shows the performance of Stage I when alternatively removing the center loss (No Center Loss) from Equation (\ref{eq:centerloss}) ($\lambda_{1,2}=0$) and the anchor image (No Anchor) when training $R_1$, thus passing from relative rotation to the more standard absolute rotation. In both cases, the performance significantly drops compared to our full method, but still outperforms the multi-binary classifier of STA. 

\begin{table}[t!]
\centering
\caption{Ablation Analysis on Stage I and Stage II}
\resizebox{0.95\textwidth}{!}{
\begin{tabular}{l@{~} c c c c c c c}

\hline
\multicolumn{8}{c}{\textbf{Ablation Study}}\\
\hline

\multicolumn{1}{c}{\multirow{1}{*}{\textbf{STAGE I} (AUC-ROC)}} & \multicolumn{1}{c|}{A $\rightarrow$ W } 
 & \multicolumn{1}{c|}{A $\rightarrow$ D } 
 & \multicolumn{1}{c|}{D $\rightarrow$ W }  
 & \multicolumn{1}{c|}{W $\rightarrow$ D} 
 & \multicolumn{1}{c|}{D $\rightarrow$ A } 
 & \multicolumn{1}{c|}{W $\rightarrow$ A } 
 & \textbf{Avg.} \\
 
\hline

\textbf{ROS} & \multicolumn{1}{c|}{90.1} & \multicolumn{1}{c|}{88.1} & \multicolumn{1}{c|}{99.4} & \multicolumn{1}{c|}{99.9} & \multicolumn{1}{c|}{87.5} & \multicolumn{1}{c|}{83.8} &\textbf{91.5}\\
Multi-Binary (from STA \cite{liu2019separate}) & \multicolumn{1}{c|}{83.2} & \multicolumn{1}{c|}{84.1} & \multicolumn{1}{c|}{86.8} & \multicolumn{1}{c|}{72.0} & \multicolumn{1}{c|}{75.7} & \multicolumn{1}{c|}{78.3} & 79.9\\
ROS - No Center loss & \multicolumn{1}{c|}{88.8} & \multicolumn{1}{c|}{83.2} & \multicolumn{1}{c|}{98.8} & \multicolumn{1}{c|}{99.8} & \multicolumn{1}{c|}{84.7} & \multicolumn{1}{c|}{84.5} & 89.9\\
ROS - No Anchor & \multicolumn{1}{c|}{84.5} & \multicolumn{1}{c|}{84.9} & \multicolumn{1}{c|}{99.1} & \multicolumn{1}{c|}{99.9} & \multicolumn{1}{c|}{87.6} & \multicolumn{1}{c|}{86.2} & 90.4\\
ROS - No Rot. Score & \multicolumn{1}{c|}{86.3} & \multicolumn{1}{c|}{82.7} & \multicolumn{1}{c|}{99.5} & \multicolumn{1}{c|}{99.9} & \multicolumn{1}{c|}{86.3} & \multicolumn{1}{c|}{82.9} & 89.6\\
ROS - No Ent. Score & \multicolumn{1}{c|}{80.7} & \multicolumn{1}{c|}{78.7} & \multicolumn{1}{c|}{99.7} & \multicolumn{1}{c|}{99.9} & \multicolumn{1}{c|}{86.6} & \multicolumn{1}{c|}{84.4} & 88.3\\
ROS - No Center loss, No Anchor & \multicolumn{1}{c|}{76.5} & \multicolumn{1}{c|}{79.1} & \multicolumn{1}{c|}{98.3} & \multicolumn{1}{c|}{99.7} & \multicolumn{1}{c|}{85.2} & \multicolumn{1}{c|}{83.5} & 87.1\\
ROS - No Rot. Score, No Anchor & \multicolumn{1}{c|}{83.9} & \multicolumn{1}{c|}{84.6} & \multicolumn{1}{c|}{99.4} & \multicolumn{1}{c|}{99.9} & \multicolumn{1}{c|}{84.7} & \multicolumn{1}{c|}{84.9} & 89.6\\
ROS - No Ent. Score, No Anchor & \multicolumn{1}{c|}{80.1} & \multicolumn{1}{c|}{81.0} & \multicolumn{1}{c|}{99.5} & \multicolumn{1}{c|}{99.7} & \multicolumn{1}{c|}{84.3} & \multicolumn{1}{c|}{83.3} & 87.9\\
ROS - No Rot. Score, No Center loss & \multicolumn{1}{c|}{80.9} & \multicolumn{1}{c|}{81.6} & \multicolumn{1}{c|}{98.9} & \multicolumn{1}{c|}{99.8} & \multicolumn{1}{c|}{85.6} & \multicolumn{1}{c|}{83.3} & 88.3\\
ROS - No Ent. Score, No Center loss & \multicolumn{1}{c|}{76.4} & \multicolumn{1}{c|}{79.8} & \multicolumn{1}{c|}{99.0} & \multicolumn{1}{c|}{98.3} & \multicolumn{1}{c|}{84.4} & \multicolumn{1}{c|}{84.3} & 87.0\\
ROS - No Ent. Score, No Center loss, No Anchor & \multicolumn{1}{c|}{78.6} & \multicolumn{1}{c|}{80.4} & \multicolumn{1}{c|}{99.0} & \multicolumn{1}{c|}{98.9} & \multicolumn{1}{c|}{86.2} & \multicolumn{1}{c|}{83.2} & 87.7\\
ROS - No Rot. Score, No Center loss, No Anchor & \multicolumn{1}{c|}{78.7} & \multicolumn{1}{c|}{82.2} & \multicolumn{1}{c|}{98.3} & \multicolumn{1}{c|}{99.8} & \multicolumn{1}{c|}{85.0} & \multicolumn{1}{c|}{82.6} & 87.8\\

\hline\hline

\multicolumn{1}{c}{\multirow{1}{*}{\textbf{STAGE II} (HOS)}}  & \multicolumn{1}{c|}{A $\rightarrow$ W } & \multicolumn{1}{c|}{A $\rightarrow$ D } & \multicolumn{1}{c|}{D $\rightarrow$ W }  & \multicolumn{1}{c|}{W $\rightarrow$ D} & \multicolumn{1}{c|}{D $\rightarrow$ A } & \multicolumn{1}{c|}{W $\rightarrow$ A } & \multicolumn{1}{c}{\textbf{Avg.}} 
 \\
\hline
\textbf{ROS}  & \multicolumn{1}{c|}{82.1}  & \multicolumn{1}{c|}{82.4}  & \multicolumn{1}{c|}{96.0}  & \multicolumn{1}{c|}{99.7}   & \multicolumn{1}{c|}{77.9}  & \multicolumn{1}{c|}{77.2}  & \textbf{85.9}   \\

ROS Stage I - GRL \cite{ganin2016domain} Stage II  & \multicolumn{1}{c|}{83.5}  & \multicolumn{1}{c|}{80.9}   & \multicolumn{1}{c|}{97.1}  & \multicolumn{1}{c|}{99.4} & \multicolumn{1}{c|}{77.3} & \multicolumn{1}{c|}{72.6} &  85.1\\

ROS Stage I - No Anchor in Stage II  & \multicolumn{1}{c|}{80.0}   & \multicolumn{1}{c|}{82.3}  & \multicolumn{1}{c|}{94.5}   & \multicolumn{1}{c|}{99.2}   & \multicolumn{1}{c|}{76.9}  & \multicolumn{1}{c|}{76.6}  & 84.9\\

ROS Stage I - No Anchor, No Entropy in Stage II  & \multicolumn{1}{c|}{80.1}   & \multicolumn{1}{c|}{84.4}  & \multicolumn{1}{c|}{97.0}   & \multicolumn{1}{c|}{99.2}   & \multicolumn{1}{c|}{76.5}  & \multicolumn{1}{c|}{72.9}  & 85.0\\

\hline
\end{tabular}
}
    \label{tab:ablationstep1} 
\end{table}

\paragraph{Why is the normality score defined the way it is?}
As defined in Equation (\ref{eq:normalityscore}), our normality score is a function of the rotation score and entropy score. The rotation score is based on the ability of $R_1$ to predict the rotation of the target samples, while the entropy score is based on the confidence of such predictions. Table~\ref{tab:ablationstep1} shows the results of Stage I when alternatively discarding either the rotation score (No Rot. Score) or the information of the entropy score (No Ent. Score). In both cases the AUC-ROC significantly decreases  compared to the full version, justifying our choice.

\paragraph{Is rotation recognition effective for domain alignment in OSDA?}
While rotation classification has already been used for CSDA~\cite{xu2019self-supervised}, its application in OSDA, where the shared target distribution could be noisy (\ie contain unknown samples) has not been studied. On the other hand, GRL~\cite{ganin2016domain} is used, under different forms, by all existing OSDA methods. We compare rotation recognition and GRL in this context by evaluating the performance of our Stage II when replacing the $R_2$ with a domain discriminator. Table~\ref{tab:ablationstep1} shows that rotation recognition performs on par with GRL, if not slightly better. Moreover we also evaluate the role of the relative rotation in the Stage II: the results in the last row of Table~\ref{tab:ablationstep1} confirm that it improves over the standard absolute rotation (No Anchor in Stage II) even when the rotation classifier is used as cross-domain adaptation strategy. Finally, the cosine distance between the source and the target domain without adaptation in Stage II ($0.188$) and with our full method ($0.109$) confirms that rotation recognition is indeed helpful to reduce the domain gap.
\begin{figure}[t!]
\centering
\includegraphics[width=0.28\linewidth]{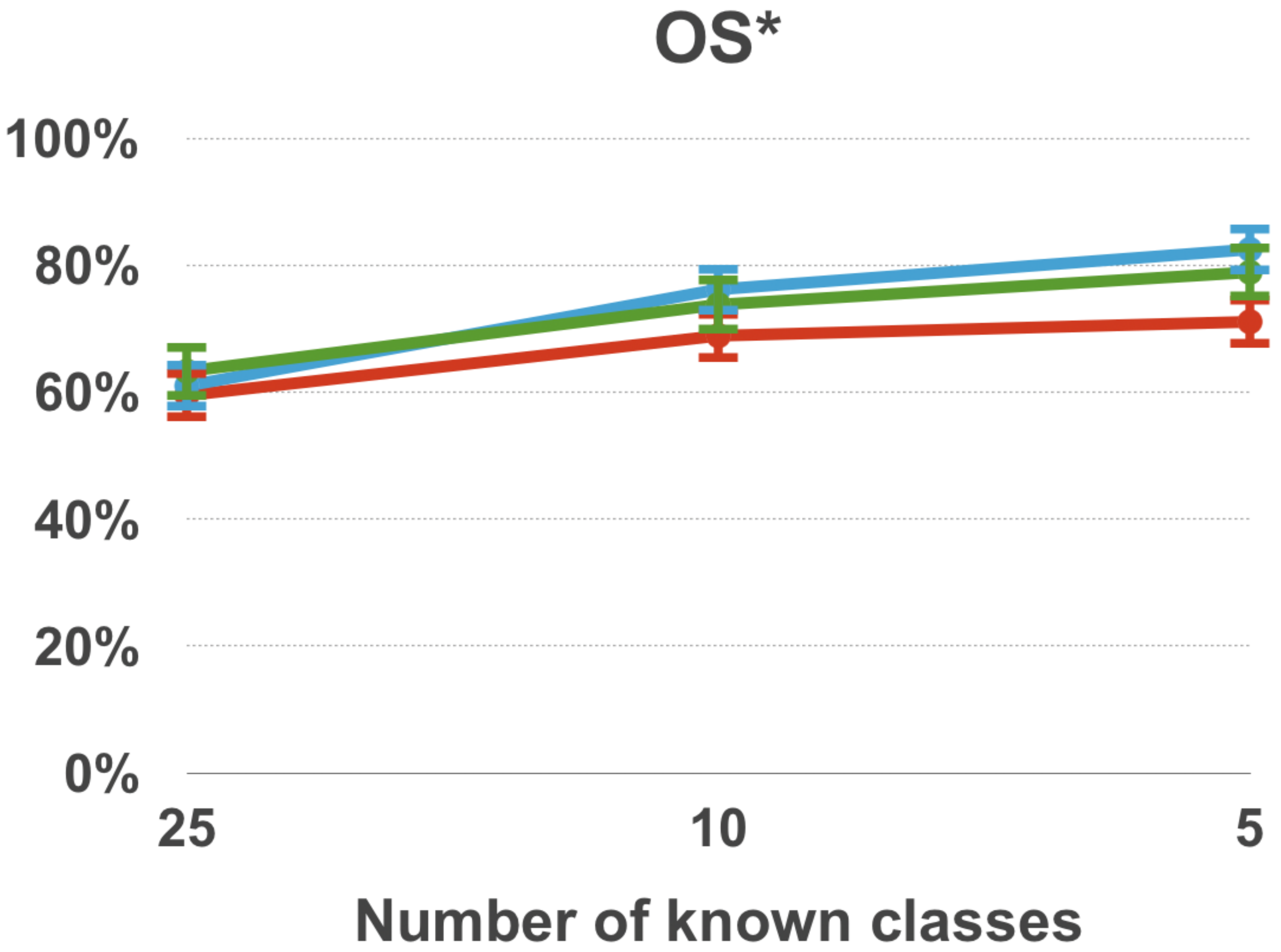} 
\includegraphics[width=0.28\linewidth]{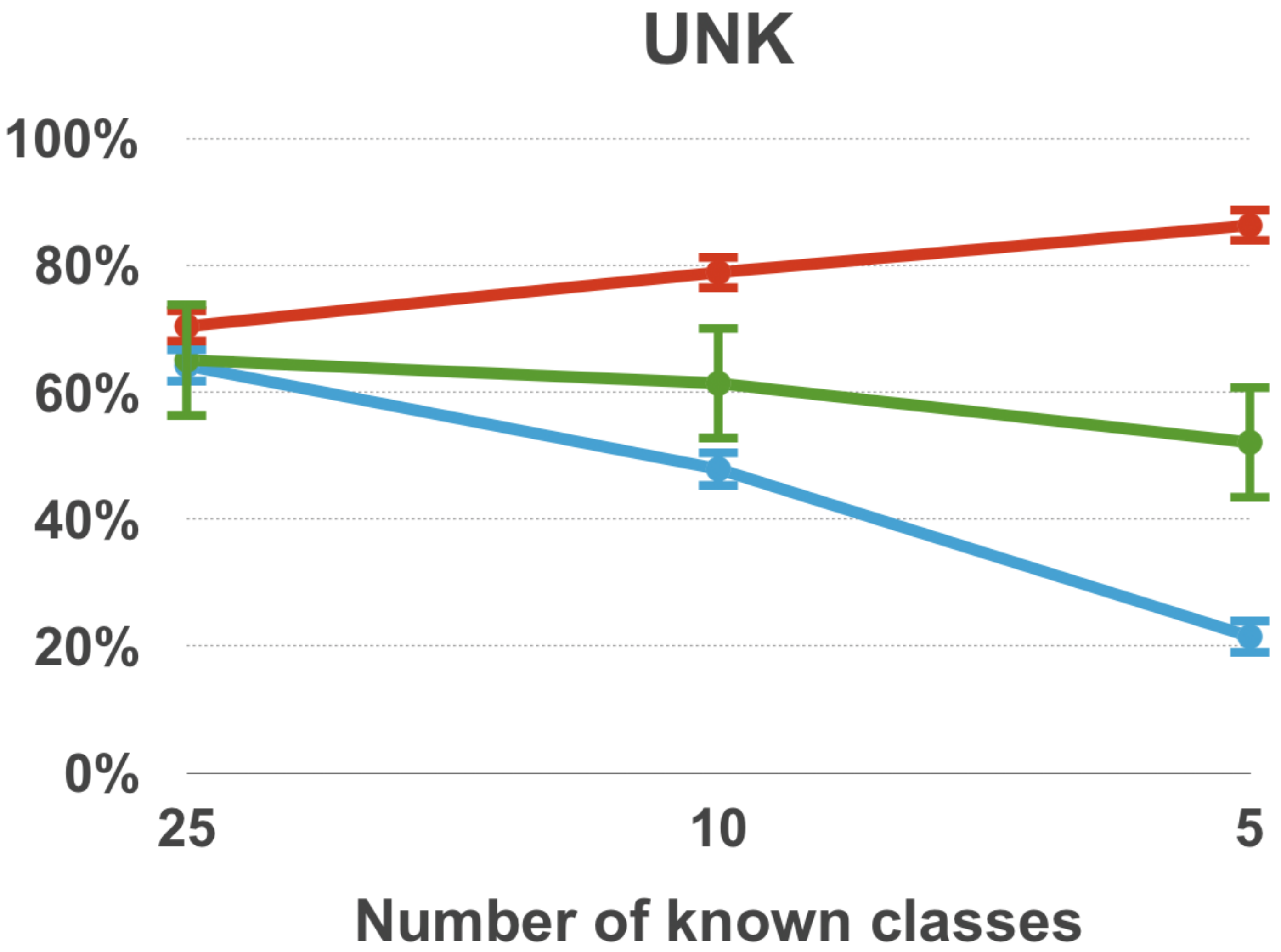} 
\includegraphics[width=0.28\linewidth]{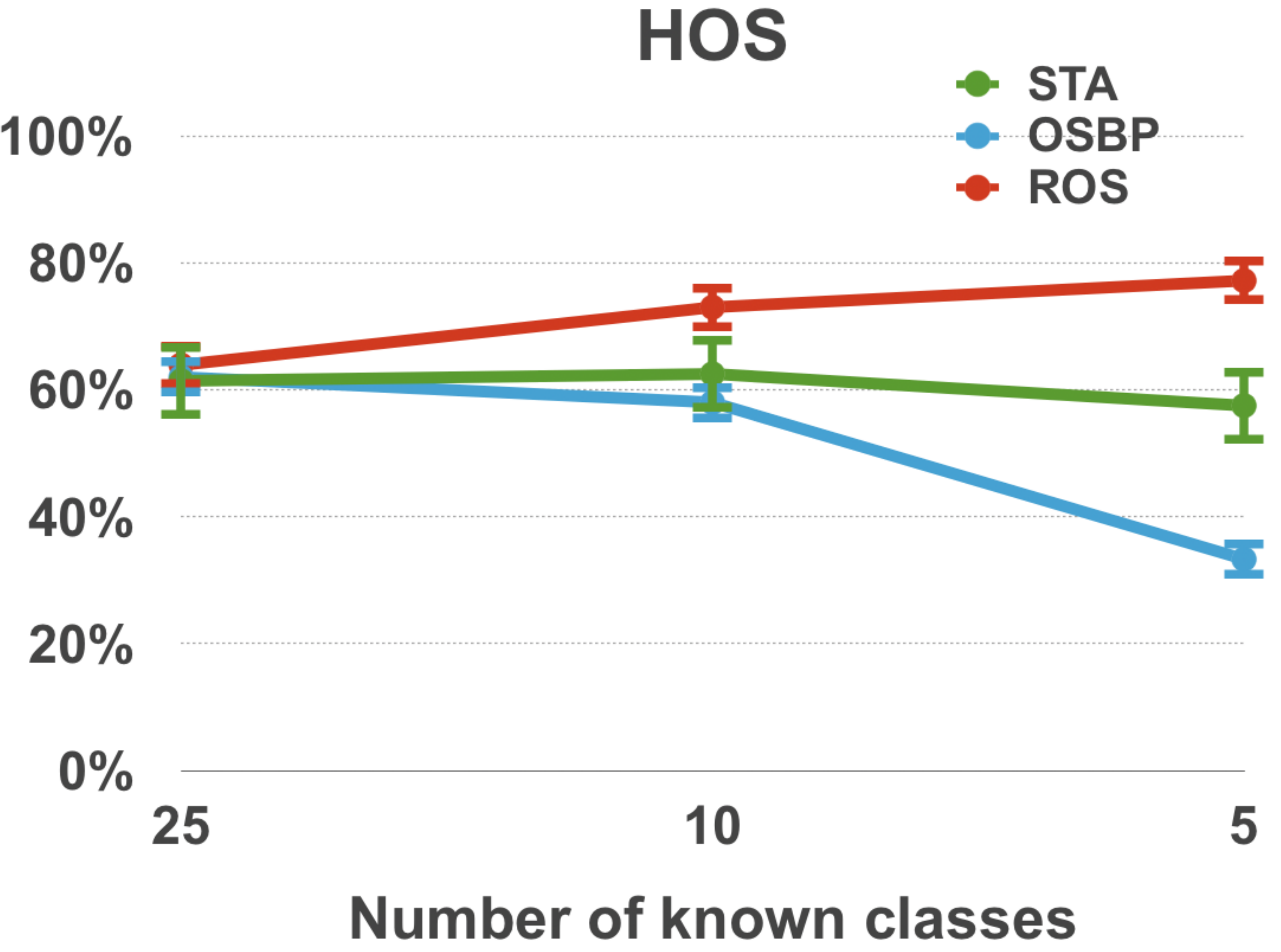}
\caption{Accuracy (\%) averaged over the three openness configurations.}
\label{fig:decreaseknownclass}
\end{figure}
\paragraph{Is our method effective on problems with a high degree of openness?}
The standard open set setting adopted in so far, presents a relatively balanced number of shared and private target classes with openness close to $0.5$. Specifically it is  $\bbO=1-\frac{10}{21}=0.52$ for Office-31 and $\bbO=1-\frac{25}{65}=0.62$ for Office-Home. In real-world problems, we can expect the number of unknown target classes to largely exceed the number of known classes, with openness approaching $1$. We investigate this setting using Office-Home and, starting from the classes sorted with ID from 0 to 64 in alphabetic order, we define the following settings with increasing openness: 
\textbf{25} known classes $\bbO=0.62$, ID:\{0-24, 25-49, 40-64\},
\textbf{10} known classes $\bbO=0.85$, ID:\{0-9, 10-19, 20-29\},
\textbf{5} known classes $\bbO=0.92$, ID:\{0-4, 5-9, 10-14\}.
Figure~\ref{fig:decreaseknownclass} shows that the performance of our best competitors, STA and OSBP, deteriorates with larger $\bbO$ due to their inability to recognize the unknown samples. On the other hand, ROS maintains a consistent performance.

\section{Discussion and conclusions}
\label{sec:conclusions}
In this paper, we present ROS: a novel method that tackles OSDA by using the self-supervised task of predicting image rotation. We show that, with simple variations of the rotation prediction task, we can first separate the target samples into known and unknown, and then align the target samples predicted as known with the source samples. Additionally, we propose HOS: a new OSDA metric defined as the harmonic mean between the accuracy of recognizing the known classes and rejecting the unknown samples. HOS overcomes the drawbacks of the current metric OS where the contribution of the unknown classes vanishes with increasing number of known classes. 

We evaluate the perfomance of ROS
on the standard Office-31 and Office-Home benchmarks, showing that it
outperforms the competing methods.
In addition, when tested on
settings with increasing openness, ROS is the only method that maintains a steady performance. HOS reveals to be crucial in this evaluation to correctly assess the performance of the methods on both known and unknown samples.
Finally, the failure in reproducing the reported results of existing methods exposes an important issue in OSDA that echoes the current reproducibility crisis in machine learning. We hope that our contributions can help laying a more solid foundation for the field.\\

\noindent\textbf{Acknowledgements} This work was partially founded by the ERC grant 637076 RoboExNovo (SB), by the H2020 ACROSSING project grant 676157 (MRL) and took advantage of the NVIDIA GPU Academic Hardware Grant (TT).

\clearpage

\section*{Appendix}
\appendix

\section{Implementation Details}
In this section we provide all the implementation details about our method ROS and the parameters used in running all our experiments.
We ran ROS on Office-31 \cite{saenko2010adapting}, Office-Home \cite{venkateswara2017deep} and using two different backbones, 
ResNet-50  \cite{he2016deep} and VGGNet \cite{simonyan2014very}. For an overall scheme of our architecture, we refer the reader to Figure 1 of the main paper.

\paragraph{Encoder $E$, ResNet-50}: it is composed by all the layers of a standard ResNet-50 up to the average pooling layer. We start from the encoder model pre-trained on ImageNet \cite{deng2009imagenet} and we update only the last convolutional block, finetuning it with learning rate $0.0003$.
\paragraph{Classifiers $C_1$, $C_2$, ResNet-50}: they are both mainly composed by two Fully Connected (FC) layers. Specifically the first  FC has output 256 and is followed by a Batch Normalization \cite{ioffe2015batch} layer and Leakly-ReLU (with negative slope angle as $0.2$). The second FC changes depending on the classifier: for $C_1$ it has $|\sC_s|$ outputs, while for $C_2$ it has $|\sC_s|+1$ outputs including the unknown category. All the layers are learned from scratch with learning rate $0.003$.
\paragraph{Rotation classifiers $R_1$, $R_2$, ResNet-50}: they both have the same structure of the classifiers described above. The only difference is in the number of outputs which is $4\times|\sC_s|$ for $R_1$ and $4$ for $R_2$. All the layers are learned from scratch with learning rate $0.003$.
\paragraph{Stage I and Stage II, ResNet-50}: The network trained in Stage I is used as starting point for Stage II, and we know that for the semantic classifier the set of categories increases by one. To take it into consideration, in Stage II we set the learning rate of the new unknown class to twice that of the known classes (already learned in Stage I).
\begin{table}[t]
\centering
\caption{Office-Home Resnet50}
\resizebox{\textwidth}{!}{
\begin{tabular}{l@{~~~}cc cccc cccc cccc  }
\hline
            \multicolumn{15}{c}{\textbf{Office-Home}}        \\
\hline
 & &  & & & \multicolumn{4}{c|}{Pr $\rightarrow$ Rw } & \multicolumn{4}{c}{Pr $\rightarrow$ Cl }  & 
 \\
        & & & & & OS & OS* & UNK & \multicolumn{1}{c|}{\textbf{\underline{HOS}}} & OS & OS* & UNK &  \multicolumn{1}{c}{\textbf{\underline{HOS}}} \\
\hline
\multicolumn{1}{c}{STA\textsubscript{sum}}& \multirow{2}{*}{\cite{liu2019separate}}  & &  & & $77.5\pm$2.3 & $78.1\pm$2.2 & $63.3\pm$8.9 & \multicolumn{1}{c|}{$69.7\pm$5.9} &  $45.7\pm$2.1 & $44.7\pm$1.9 & $71.5\pm$6.6 & \multicolumn{1}{c}{$55.0\pm$3.3} \\

\multicolumn{1}{c}{STA\textsubscript{max}} & & & & & $75.7\pm$2.4 & $76.2\pm$2.2 & $64.3\pm$9.8 & \multicolumn{1}{c|}{$69.5\pm$6.2} & $45.1\pm$4.5 & $44.2\pm$4.7 & $67.1\pm$1.9 & \multicolumn{1}{c}{$53.2\pm$2.8}  \\

\multicolumn{1}{c}{OSBP}\cite{saito2018open} & & & & & $76.0\pm$1.3 & $76.2\pm$1.3 & $71.7\pm$1.6 & \multicolumn{1}{c|}{$73.9$$\pm$1.4} & $45.3\pm$1.1 & 44.5$\pm$1.2 & 66.3$\pm$1.8 & \multicolumn{1}{c}{$53.2$$\pm$0.8}   \\

\multicolumn{1}{c}{UAN}\cite{you2019universal} & &  &  & & 81.0$\pm$0.1 & 84.0$\pm$0.1 & 0.1$\pm$0.0 & \multicolumn{1}{c|}{0.2$\pm$0.0} & 57.1$\pm$0.2  & 59.1$\pm$0.8 & 0.0$\pm$0.0 & \multicolumn{1}{c}{0.0$\pm$0.0}  \\
\multicolumn{1}{c}{\textbf{ROS}} &  & & & & 71.1$\pm$1.3 & 70.8$\pm$1.4 & 78.4$\pm$1.8 &  \multicolumn{1}{c|}{\textbf{74.4$\pm$0.1}} & 47.5$\pm$0.9 & 46.5$\pm$1.0 & 71.2$\pm$0.9 & \multicolumn{1}{c}{\textbf{56.3$\pm$0.5}}  \\

\hline

  & &  & & &  \multicolumn{4}{c|}{Pr $\rightarrow$ Ar }  &  \multicolumn{4}{c}{Ar $\rightarrow$ Pr }  & 

 \\
        & & & & & OS & OS* & UNK & \multicolumn{1}{c|}{\textbf{\underline{HOS}}} & OS & OS* & UNK &  \multicolumn{1}{c}{\textbf{\underline{HOS}}} \\
\hline
\multicolumn{1}{c}{STA\textsubscript{sum}}& &  & & &  $56.1\pm$2.8 & $55.4\pm$2.5 & $73.7\pm$12.0 & \multicolumn{1}{c|}{$63.1\pm$5.6}  &   $68.4\pm$3.4 & $68.7\pm$3.8 & $59.7\pm$5.5 & \multicolumn{1}{c}{$63.7\pm$1.4} \\
\multicolumn{1}{c}{STA\textsubscript{max}} & & & & &  $54.9\pm$6.3 & $54.2\pm$6.4 & $72.4\pm$7.6 & \multicolumn{1}{c|}{$61.9\pm$6.3} & $67.3\pm$1.0 & $68.0\pm$1.8 & $48.4\pm$22.6 &
\multicolumn{1}{c}{$54.0\pm$17.5}\\
\multicolumn{1}{c}{OSBP} & & & & & $59.4\pm$0.7 & $59.1\pm$0.8 & $68.1\pm$0.8 & \multicolumn{1}{c|}{\textbf{63.2$\pm$0.2}}  & $71.3\pm$0.5 & $71.8\pm$0.5 & $59.8\pm$0.5 & \multicolumn{1}{c}{$65.2\pm$0.4} \\
\multicolumn{1}{c}{UAN} & & & &  & 78.1$\pm$0.1  & 81.1$\pm$0.2 & 0.0$\pm$0.0 & \multicolumn{1}{c|}{0.0$\pm$0.0} & 78.1$\pm$0.1  & 81.1$\pm$0.2 & 0.0$\pm$0.0 & \multicolumn{1}{c}{0.0$\pm$0.0} \\
\multicolumn{1}{c}{\textbf{ROS}} & & &  & & 57.6$\pm$0.8 & 57.3$\pm$0.8 & 64.3$\pm$1.7 & \multicolumn{1}{c|}{60.6$\pm$1.2}  & 68.4$\pm$1.1 & 68.4$\pm$1.2 & 70.3$\pm$1.6 & \multicolumn{1}{c}{\textbf{69.3$\pm$0.2}} \\
\hline

  & &  & & &  \multicolumn{4}{c|}{Ar $\rightarrow$ Rw }  &  \multicolumn{4}{c}{Ar $\rightarrow$ Cl }  & 
 \\
        & & & & & OS & OS* & UNK & \multicolumn{1}{c|}{\textbf{\underline{HOS}}} & OS & OS* & UNK &  \multicolumn{1}{c}{\textbf{\underline{HOS}}}  \\
\hline

\multicolumn{1}{c}{STA\textsubscript{sum}}& &  &  & & $80.0\pm$0.6 & $81.1\pm$0.4 & $50.5\pm$6.3 & \multicolumn{1}{c|}{$62.1$$\pm$5.0}   & $51.3\pm$2.5 & $50.8\pm$2.7 & $63.4\pm$3.4 & \multicolumn{1}{c}{$56.3$$\pm$1.2}  \\
\multicolumn{1}{c}{STA\textsubscript{max}} & & & & & $77.9\pm$0.4 & $78.6\pm$0.4 & 60.4$\pm$1.9 & \multicolumn{1}{c|}{$68.3$$\pm$1.2} & $47.0\pm$5.8 & 46.0$\pm$6.3 & 72.3$\pm$6.2 & \multicolumn{1}{c}{$55.8$$\pm$2.6}\\
\multicolumn{1}{c}{OSBP} & & & & & $78.8\pm$0.8 & $79.3\pm$0.9 & $67.5\pm$0.3 & \multicolumn{1}{c|}{$72.9$$\pm$0.5} &   50.7$\pm$0.7 & 50.2$\pm$0.9 & 61.1$\pm$0.7 & \multicolumn{1}{c}{$55.1$$\pm$0.4} \\
\multicolumn{1}{c}{UAN} & & &  & & 85.1$\pm$0.1  & 88.2$\pm$0.2 & 0.1$\pm$0.0 & \multicolumn{1}{c|}{0.2$\pm$0.0} &  60.0$\pm$0.1 & 62.4$\pm$0.1 & 0.0$\pm$0.0 & \multicolumn{1}{c}{0.0$\pm$0.0} \\
\multicolumn{1}{c}{\textbf{ROS}} & & &  & & 75.9$\pm$1.1 & 75.8$\pm$1.2 & 77.2$\pm$0.9 & \multicolumn{1}{c|}{\textbf{76.5$\pm$0.4}}& 51.5$\pm$1.0 & 50.6$\pm$1.1 & 74.1$\pm$1.0 & \multicolumn{1}{c}{\textbf{60.1$\pm$0.4}}\\

\hline

  & &  & & &  \multicolumn{4}{c|}{Rw $\rightarrow$ Ar }  &  \multicolumn{4}{c}{Rw $\rightarrow$ Pr }  & 
 \\
        & & & & & OS & OS* & UNK & \multicolumn{1}{c|}{\textbf{\underline{HOS}}} & OS & OS* & UNK &  \multicolumn{1}{c}{\textbf{\underline{HOS}}}  \\
\hline

\multicolumn{1}{c}{STA\textsubscript{sum}}& &  &  & &  67.7$\pm$2.8 & 67.9$\pm$2.8 & 62.3$\pm$2.8 & \multicolumn{1}{c|}{65.0$\pm$2.8}  & 77.1$\pm$1.7 & 77.9$\pm$1.7 & 58.0$\pm$4.5 & \multicolumn{1}{c}{66.4$\pm$3.3} \\
\multicolumn{1}{c}{STA\textsubscript{max}} & &  &  & &  67.5$\pm$1.7 & 67.5$\pm$1.8 & 66.7$\pm$3.0 & \multicolumn{1}{c|}{67.1$\pm$1.2} & 76.3$\pm$0.4 & 77.1$\pm$0.5 & 55.4$\pm$1.5 & \multicolumn{1}{c}{64.5$\pm$1.0}\\
\multicolumn{1}{c}{OSBP}& &  &  & & 66.1$\pm$0.5 & 66.1$\pm$0.6 & 67.3$\pm$0.6 & \multicolumn{1}{c|}{66.7$\pm$0.4} & 76.0$\pm$0.7 & 76.3$\pm$0.7 & 68.6$\pm$2.1 & \multicolumn{1}{c}{72.3$\pm$1.3} \\
\multicolumn{1}{c}{UAN} & &  &  & &  74.8$\pm$0.0 & 77.5$\pm$0.1 & 0.1$\pm$0.0 & \multicolumn{1}{c|}{0.2$\pm$0.0} & 82.1$\pm$0.1  & 85.0$\pm$0.2 & 0.1$\pm$0.1 & \multicolumn{1}{c}{0.1$\pm$0.1}\\
\multicolumn{1}{c}{\textbf{ROS}}& &  &  & &   67.1$\pm$1.0 & 67.0$\pm$1.1 & 70.8$\pm$2.0 & \multicolumn{1}{c|}{\textbf{68.8$\pm$0.6}} & 72.2$\pm$0.7 & 72.0$\pm$0.6 & 80.0$\pm$1.1 & \multicolumn{1}{c}{\textbf{75.7$\pm$0.9}}\\

\hline

  & &  & & &  \multicolumn{4}{c|}{Rw $\rightarrow$ Cl }  &  \multicolumn{4}{c}{Cl $\rightarrow$ Rw }  & 
   \\
        & & & & & OS & OS* & UNK & \multicolumn{1}{c|}{\textbf{\underline{HOS}}} & OS & OS* & UNK &  \multicolumn{1}{c}{\textbf{\underline{HOS}}}   \\
\hline

\multicolumn{1}{c}{STA\textsubscript{sum}}& &  &  & &  51.7$\pm$3.0 & 51.4$\pm$3.3 & 57.9$\pm$5.3 & \multicolumn{1}{c|}{54.2$\pm$0.9}   & 69.5$\pm$2.7 & 69.8$\pm$2.4 & 63.2$\pm$8.9 & \multicolumn{1}{c}{66.3$\pm$5.9} \\
\multicolumn{1}{c}{STA\textsubscript{max}} & &  &  & &   50.4$\pm$2.5 & 49.9$\pm$2.9 & 61.1$\pm$9.8 & \multicolumn{1}{c|}{54.5$\pm$2.6} & 67.0$\pm$3.0 & 67.0$\pm$2.8 & 66.7$\pm$8.7 & \multicolumn{1}{c}{66.8$\pm$5.7} \\
\multicolumn{1}{c}{OSBP}& &  &  & &  48.5$\pm$0.5 & 48.0$\pm$0.5 & 63.0$\pm$0.6 & \multicolumn{1}{c|}{54.5$\pm$0.2} & 71.9$\pm$0.9 & 72.0$\pm$0.9 & 69.2$\pm$0.2 & \multicolumn{1}{c}{\textbf{70.6$\pm$0.4}} \\
\multicolumn{1}{c}{UAN} & &  &  & &  63.5$\pm$0.1 & 66.2$\pm$0.5 & 0.0$\pm$0.0 & \multicolumn{1}{c|}{0.0$\pm$0.0} & 77.7$\pm$0.4  & 80.6$\pm$0.4 & 0.1$\pm$0.0 & \multicolumn{1}{c}{0.2$\pm$0.0}\\
\multicolumn{1}{c}{\textbf{ROS}}& &  &  & &   52.3$\pm$0.9 & 51.5$\pm$0.9 & 73.0$\pm$0.8 & \multicolumn{1}{c|}{\textbf{60.4$\pm$0.5}} & 65.6$\pm$0.3 & 65.3$\pm$0.3 & 72.2$\pm$1.6 & \multicolumn{1}{c}{68.6$\pm$0.7}\\

\hline

  & & &  \multicolumn{4}{c|}{Cl $\rightarrow$ Ar }  &  \multicolumn{4}{c|}{Cl $\rightarrow$ Pr }  & \multicolumn{4}{c}{Avg.} 
   \\
        &   & & OS & OS* & UNK & \multicolumn{1}{c|}{\textbf{\underline{HOS}}} & OS & OS* & UNK &  \multicolumn{1}{c|}{\textbf{\underline{HOS}}} & OS & OS* & UNK &  \multicolumn{1}{c}{\textbf{\underline{HOS}}} \\
\hline

\multicolumn{1}{c}{STA\textsubscript{sum}}& &  &   53.5$\pm$3.0 & 53.0$\pm$3.1 & 63.9$\pm$1.2 & \multicolumn{1}{c|}{57.9$\pm$2.3}  & 61.5$\pm$2.4 & 61.4$\pm$2.4 & 63.5$\pm$3.3 & \multicolumn{1}{c|}{62.5$\pm$2.8} & 63.3$\pm$2.1 & 63.4$\pm$2.1 & 62.6$\pm$2.3 & \multicolumn{1}{c}{61.9$\pm$2.1}  \\
\multicolumn{1}{c}{STA\textsubscript{max}} & & &  52.0$\pm$2.2 & 51.4$\pm$2.3 & 65.0$\pm$2.2 & \multicolumn{1}{c|}{57.4$\pm$0.7}  & 61.7$\pm$1.6 & 61.8$\pm$1.7 & 59.1$\pm$1.1 & \multicolumn{1}{c|}{60.4$\pm$0.7} & 61.9$\pm$2.4 & 61.8$\pm$2.6 & 63.3$\pm$1.9 & \multicolumn{1}{c}{61.1$\pm$0.3}\\
\multicolumn{1}{c}{OSBP}  &  & &  59.8$\pm$0.4 & 59.4$\pm$0.5 & 70.3$\pm$1.3 & \multicolumn{1}{c|}{\textbf{64.3$\pm$0.4}}  & 66.9$\pm$1.5 & 67.0$\pm$1.6 & 62.7$\pm$2.3 & \multicolumn{1}{c|}{64.7$\pm$0.7}  & 64.2$\pm$0.1 & 64.1$\pm$0.1 & 66.3$\pm$0.4 & \multicolumn{1}{c}{64.7$\pm$0.2}\\
\multicolumn{1}{c}{UAN}   &  & &  67.8$\pm$0.3  & 70.5$\pm$0.5 & 0.0$\pm$0.0 & \multicolumn{1}{c|}{0.0$\pm$0.0} & 71.3$\pm$0.1 & 74.0$\pm$0.2 & 0.1$\pm$0.0 & \multicolumn{1}{c|}{0.2$\pm$0.0} &  72.5$\pm$0.0 & 75.2$\pm$0.1 & 0.0$\pm$0.0 & \multicolumn{1}{c}{0.1$\pm$0.0} \\
\multicolumn{1}{c}{\textbf{ROS}}  &  & &  54.1$\pm$1.0 & 53.6$\pm$1.1 & 65.5$\pm$2.8 & \multicolumn{1}{c|}{58.9$\pm$0.5} & 60.3$\pm$0.4 & 59.8$\pm$0.4 & 71.6$\pm$1.1 & \multicolumn{1}{c|}{\textbf{65.2$\pm$0.7}} & 62.0$\pm$0.2 & 61.6$\pm$0.2 & 72.4$\pm$0.8 & \multicolumn{1}{c}{\textbf{66.2$\pm$0.3}} \\

\hline

\end{tabular}
}

    \label{tab:officehomeResnet50sup} 
\end{table}
\begin{table}[t]
\centering
\caption{Office-31}
\resizebox{\textwidth}{!}{
\begin{tabular}{l@{~~~~~}cc cccc cccc cccc  }
\hline

&       \multicolumn{12}{c}{\textbf{ Office-31 ResNet-50}}        \\
\hline
& & & & \multicolumn{4}{c|}{A $\rightarrow$ W } & \multicolumn{4}{c}{A $\rightarrow$ D } 
 \\
     & & & & OS & OS* & UNK & \multicolumn{1}{c|}{\textbf{\underline{HOS}}} & OS & OS* & UNK &  \multicolumn{1}{c}{\textbf{\underline{HOS}}}\\
     
\hline
\multicolumn{1}{c}{STA\textsubscript{sum}} & \multirow{2}{*}{\cite{liu2019separate}} & & & 89.0$\pm$4.0 & 92.1$\pm$4.6 & 58.0$\pm$5.7 & \multicolumn{1}{c|}{71.0$\pm$4.0} & 90.8$\pm$2.6 & 95.4$\pm$2.8 & 45.5$\pm$1.6 & \multicolumn{1}{c}{61.6$\pm$1.7} \\
\multicolumn{1}{c}{STA\textsubscript{max}} & & & & 85.0$\pm$4.8 & 86.7$\pm$5.4 & 67.6$\pm$1.3 & \multicolumn{1}{c|}{75.9$\pm$1.3} & 88.5$\pm$2.1 & 91.0$\pm$2.6 & 63.9$\pm$3.8 & \multicolumn{1}{c}{75.0$\pm$2.2} \\
\multicolumn{1}{c}{OSBP}\cite{saito2018open} & & & & 86.0$\pm$1.1 & 86.8$\pm$1.2 & 79.2$\pm$0.4 & \multicolumn{1}{c|}{\textbf{82.7$\pm$0.6}} & 89.2$\pm$0.4 & 90.5$\pm$0.4 & 75.5$\pm$1.4 & \multicolumn{1}{c}{\textbf{82.4$\pm$0.9}}\\
\multicolumn{1}{c}{UAN}\cite{you2019universal} & & & & 89.4$\pm$0.4 & 95.5$\pm$0.1 & 31.0$\pm$0.9 &  \multicolumn{1}{c|}{46.8$\pm$1.0} & 89.5$\pm$0.4  & 95.6$\pm$0.5  & 24.4$\pm$0.9 & \multicolumn{1}{c}{38.9$\pm$1.1} \\
\multicolumn{1}{c}{\textbf{ROS}}  & & & & 87.3$\pm$1.5 & 88.4$\pm$1.7 & 76.7$\pm$2.4 & \multicolumn{1}{c|}{82.1$\pm$1.4} & 86.7$\pm$0.5 & 87.5$\pm$0.6 & 77.8$\pm$0.6 & \multicolumn{1}{c}{\textbf{82.4$\pm$0.6}}\\
     
\hline
& & & & \multicolumn{4}{c|}{D $\rightarrow$ W} & \multicolumn{4}{c}{W $\rightarrow$ D } 
 \\
 
      &  & & & OS & OS* & UNK & \multicolumn{1}{c|}{\textbf{\underline{HOS}}} & OS & OS* & UNK &  \multicolumn{1}{c}{\textbf{\underline{HOS}}}\\
\hline
\multicolumn{1}{c}{STA\textsubscript{sum}}& & & &  92.8$\pm$1.3 & 97.1$\pm$0.8 & 49.7$\pm$8.1 & \multicolumn{1}{c|}{65.5$\pm$7.0} & 92.2$\pm$0.5 & 96.6$\pm$0.4 & 48.5$\pm$6.0 & \multicolumn{1}{c}{64.4$\pm$5.1} \\
\multicolumn{1}{c}{STA\textsubscript{max}}& & & & 90.6$\pm$2.8 & 94.1$\pm$3.2 & 55.5$\pm$1.3 & \multicolumn{1}{c|}{69.8$\pm$0.2} & 83.4$\pm$6.4 & 84.9$\pm$7.2 & 67.8$\pm$5.0 & \multicolumn{1}{c}{75.2$\pm$3.6}\\
\multicolumn{1}{c}{OSBP}& & & & 97.5$\pm$0.5 & 97.7$\pm$0.2 & 96.7$\pm$2.7 & \multicolumn{1}{c|}{\textbf{97.2$\pm$1.4}} & 97.8$\pm$1.1 & 99.1$\pm$1.0 & 84.2$\pm$2.2 & \multicolumn{1}{c}{91.1$\pm$1.6}\\
\multicolumn{1}{c}{UAN}& & & &  95.5$\pm$0.1 & 99.8$\pm$0.0 & 52.5$\pm$1.1 & \multicolumn{1}{c|}{68.8$\pm$1.0}&  94.7$\pm$0.4  & 81.5$\pm$32.0 & 41.4$\pm$4.2 & \multicolumn{1}{c}{53.0$\pm$9.0} \\
\multicolumn{1}{c}{\textbf{ROS}}& & & & 98.7$\pm$0.5 & 99.3$\pm$0.4 & 93.0$\pm$2.5 & \multicolumn{1}{c|}{96.0$\pm$1.5} & 99.9$\pm$0.0 & 100.0$\pm$0.0 & 99.4$\pm$0.0 & \multicolumn{1}{c}{\textbf{99.7$\pm$0.0}} \\

\hline
& & \multicolumn{4}{c|}{D $\rightarrow$ A} & \multicolumn{4}{c|}{W $\rightarrow$ A } & \multicolumn{4}{c}{Avg.} 
 \\
 
      &  & OS & OS* & UNK & \multicolumn{1}{c|}{\textbf{\underline{HOS}}} & OS & OS* & UNK &  \multicolumn{1}{c|}{\textbf{\underline{HOS}}} & OS & OS* & UNK &  \multicolumn{1}{c}{\textbf{\underline{HOS}}}\\
\hline
\multicolumn{1}{c}{STA\textsubscript{sum}} & & 90.5$\pm$1.9 & 94.1$\pm$2.1 & 55.0$\pm$1.7 & \multicolumn{1}{c|}{69.4$\pm$1.2} & 87.9$\pm$7.6 & 92.1$\pm$9.0 & 46.2$\pm$8.2 & \multicolumn{1}{c|}{60.9$\pm$5.4} & 90.6$\pm$1.8 & 94.6$\pm$2.0 & 50.5$\pm$0.8 & \multicolumn{1}{c}{65.5$\pm$0.3}    \\
\multicolumn{1}{c}{STA\textsubscript{max}} & & 81.5$\pm$5.1 & 83.1$\pm$6.2 & 65.9$\pm$5.0 & \multicolumn{1}{c|}{73.2$\pm$0.9} & 66.4$\pm$14.5 & 66.2$\pm$16.3 & 68.0$\pm$5.2 & \multicolumn{1}{c|}{66.1$\pm$7.1} & 82.6$\pm$2.1 & 84.3$\pm$2.4 & 64.8$\pm$0.9 & \multicolumn{1}{c}{72.5$\pm$0.8}\\
\multicolumn{1}{c}{OSBP} & & 75.8$\pm$0.3 & 76.1$\pm$0.4 & 72.3$\pm$1.2 & \multicolumn{1}{c|}{75.1$\pm$1.2} & 73.1$\pm$0.2 & 73.0$\pm$0.2 & 74.4$\pm$0.7 & \multicolumn{1}{c|}{73.7$\pm$0.3} & 86.6$\pm$0.1 & 87.2$\pm$0.1 & 80.4$\pm$0.7 & \multicolumn{1}{c}{83.7$\pm$0.4}\\
\multicolumn{1}{c}{UAN} & &  89.9$\pm$0.2  & 93.5$\pm$0.1 & 53.4$\pm$0.6 & \multicolumn{1}{c|}{68.0$\pm$0.5} & 89.5$\pm$0.6  & 94.1$\pm$0.2 & 38.8$\pm$0.5 & \multicolumn{1}{c|}{54.9$\pm$0.5} & 91.4$\pm$0.1  & 93.4$\pm$5.3 & 40.3$\pm$0.7 & \multicolumn{1}{c}{55.1$\pm$1.4}\\
\multicolumn{1}{c}{\textbf{ROS}} & & 75.4$\pm$0.8 & 74.8$\pm$1.0 & 81.2$\pm$0.9 & \multicolumn{1}{c|}{\textbf{77.9$\pm$0.2}} & 71.3$\pm$0.5 & 69.7$\pm$0.6 & 86.6$\pm$2.8 & \multicolumn{1}{c|}{\textbf{77.2$\pm$1.0}}  & 86.6$\pm$0.4 & 86.6$\pm$0.5 & 85.8$\pm$0.1 & \multicolumn{1}{c}{\textbf{85.9$\pm$0.2}}\\
\hline\hline
&       \multicolumn{12}{c}{\textbf{Office-31 VGGNet}}        \\
\hline
& & & & \multicolumn{4}{c|}{A $\rightarrow$ W } & \multicolumn{4}{c}{A $\rightarrow$ D } 
 \\
     & & & & OS & OS* & UNK & \multicolumn{1}{c|}{\textbf{\underline{HOS}}} & OS & OS* & UNK &  \multicolumn{1}{c}{\textbf{\underline{HOS}}}\\
     
\hline
\multicolumn{1}{c}{OSBP}\cite{saito2018open} & & & & 79.1$\pm$0.8 & 79.4$\pm$1.2 & 75.8$\pm$3.4 & \multicolumn{1}{c|}{77.5$\pm$1.2} & 86.8$\pm$6.3 & 87.9$\pm$6.4 & 75.2$\pm$6.1 & \multicolumn{1}{c}{\textbf{81.0$\pm$6.0}} \\
\multicolumn{1}{c}{\textbf{ROS}} & & & & 80.4$\pm$2.4 & 80.3$\pm$2.5 & 81.7$\pm$1.7 & \multicolumn{1}{c|}{\textbf{81.0$\pm$2.1}} & 81.3$\pm$1.0 & 81.8$\pm$1.0 & 76.5$\pm$0.6 & \multicolumn{1}{c}{79.0$\pm$0.8}\\
\hline
\multicolumn{1}{c}{\textcolor{gray}{AoD \cite{feng2019attract}}} & & & & \textcolor{gray}{86.4} & \textcolor{gray}{87.7} & \textcolor{gray}{73.4} & \multicolumn{1}{c|}{\textcolor{gray}{79.9}} & \textcolor{gray}{90.1} & \textcolor{gray}{92.0} & \textcolor{gray}{71.1} & \multicolumn{1}{c}{\textcolor{gray}{79.3}} \\
     
\hline
& & & &\multicolumn{4}{c|}{D $\rightarrow$ W} & \multicolumn{4}{c}{W $\rightarrow$ D } 
 \\
 
      &  & & & OS & OS* & UNK & \multicolumn{1}{c|}{\textbf{\underline{HOS}}} & OS & OS* & UNK &  \multicolumn{1}{c}{\textbf{\underline{HOS}}}\\
\hline

\multicolumn{1}{c}{OSBP} & & & & 96.4$\pm$0.6 & 96.8$\pm$0.7 & 93.4$\pm$0.8 & \multicolumn{1}{c|}{\textbf{95.0$\pm$0.4}} & 97.5$\pm$0.3 & 98.9$\pm$0.4 & 84.2$\pm$0.8 & \multicolumn{1}{c}{91.0$\pm$0.5} \\
\multicolumn{1}{c}{\textbf{ROS}} & & & & 98.7$\pm$0.4 & 99.5$\pm$0.4 & 89.9$\pm$0.9 & \multicolumn{1}{c|}{94.4$\pm$0.6} & 99.4$\pm$0.0 & 99.3$\pm$0.0 & 100.0$\pm$0.0 & \multicolumn{1}{c}{\textbf{99.7$\pm$0.0}}\\
\hline
\multicolumn{1}{c}{\textcolor{gray}{AoD \cite{feng2019attract}}} & & & & \textcolor{gray}{97.9} & \textcolor{gray}{99.8} & \textcolor{gray}{78.9} & \multicolumn{1}{c|}{\textcolor{gray}{88.1}} & \textcolor{gray}{98.2} & \textcolor{gray}{99.3} & \textcolor{gray}{87.2} & \multicolumn{1}{c}{\textcolor{gray}{92.9}}

\\
\hline

& & \multicolumn{4}{c|}{D $\rightarrow$ A} & \multicolumn{4}{c|}{W $\rightarrow$ A  } & \multicolumn{4}{c}{Avg. } 
 \\
 
      &  & OS & OS* & UNK & \multicolumn{1}{c|}{\textbf{\underline{HOS}}} & OS & OS* & UNK &  \multicolumn{1}{c|}{\textbf{\underline{HOS}}}  & OS & OS* & UNK &  \multicolumn{1}{c}{\textbf{\underline{HOS}}} \\
\hline

\multicolumn{1}{c}{OSBP} & & 75.1$\pm$1.3 & 74.4$\pm$1.5 & 82.4$\pm$1.2 & \multicolumn{1}{c|}{\textbf{78.2}$\pm$0.6} &
70.3$\pm$4.5 & 69.7$\pm$4.5 & 76.4$\pm$5.3 & \multicolumn{1}{c|}{72.9$\pm$4.8}  & 84.2$\pm$0.4 & 84.5$\pm$0.4 & 81.2$\pm$1.4 & \multicolumn{1}{c}{82.6$\pm$0.8}\\
\multicolumn{1}{c}{\textbf{ROS}} &  & 77.0$\pm$0.0 & 76.7$\pm$0.4  & 79.6$\pm$3.5 & \multicolumn{1}{c|}{78.1$\pm$1.4} & 64.9$\pm$0.2 & 62.2$\pm$0.2 & 91.6$\pm$0.5 & \multicolumn{1}{c|}{\textbf{74.1$\pm$0.0}} & 83.6$\pm$0.4 & 83.3$\pm$0.4 & 86.5$\pm$0.3 & \multicolumn{1}{c}{\textbf{84.4$\pm$0.2}}\\
\hline
\multicolumn{1}{c}{\textcolor{gray}{AoD \cite{feng2019attract}}} & & \textcolor{gray}{81.6} & \textcolor{gray}{88.4} & \textcolor{gray}{13.6} & \multicolumn{1}{c|}{\textcolor{gray}{23.6}} & \textcolor{gray}{80.3} & \textcolor{gray}{82.6} & \textcolor{gray}{57.3} & \multicolumn{1}{c|}{\textcolor{gray}{67.7}}  & \textcolor{gray}{89.1} & \textcolor{gray}{91.6} & \textcolor{gray}{63.6} & \multicolumn{1}{c}{\textcolor{gray}{71.9}}\\
\hline

\end{tabular} 
}

    \label{tab:office31Resnet50sup} 
\end{table}

\paragraph{Encoder $E$, VGGNet}: it is composed by all the layers of a standard VGG-19 up to the second fully connected layer. We start from the encoder model pre-trained on ImageNet \cite{deng2009imagenet} and we update only the last two FC layers, finetuning it with learning rate $0.0003$.
\paragraph{Classifiers $C_1$, $C_2$, $R_1$, $R_2$, VGGNet}: they have exactly the same structure used for the ResNet-50 case described above.
\paragraph{Stage I and Stage II, VGGNet}: The network trained in Stage I is not used as starting point for Stage II. Still we consider the learning rate of the extra unknown class in Stage II higher with respect to the other classes (1.5), but lower than the value used in case of ResNet-50 (2), where Stage II was inheriting the model of Stage I.
We also tried to inherit the Stage I model for Stage II as done in the ResNet case, but for VGG that setting produced lower results.
\paragraph{Office-31, ResNet-50}: batch size $32$, learning rate defined as specified above and decreasing during training with inverse decay scheduling. We used SGD with momentum, setting the weight decay as $0.0005$ and momentum as $0.9$. The loss weights are set as $\lambda_{1,1}= \lambda_{2,2}=3$ and $\lambda_{1,2}=\lambda_{2,1}=0.1$. We ran ROS with 80 epochs for Stage I and 80 for Stage II. Each experiment is repeated three times taking the result on the target at the last epoch.

\paragraph{Office-31, VGGNet}: batch size $32$, learning rate defined as specified above and decreasing during training with inverse decay scheduling. We used SGD with momentum, setting the weight decay as $0.0005$ and momentum as $0.9$. The loss weights are set as $\lambda_{1,1}= \lambda_{2,2}=3$ and $\lambda_{1,2}=\lambda_{2,1}=0.1$. We ran ROS with 100 epochs for Stage I and 200 for Stage II. Each experiment is repeated three times taking the result on the target at the last epoch.

\paragraph{Office-Home, ResNet-50}: batch size $32$, learning rate defined as specified above and decreasing during training with inverse decay scheduling. We used SGD with momentum, setting the weight decay as $0.0005$ and momentum as $0.9$. The loss weights are set as $\lambda_{1,1}= \lambda_{2,2}=3$ and $\lambda_{2,1}=0.1$. With respect to the previous cases, for this dataset adding the center loss to the rotation classifier $R_1$ seems less relevant: we kept it in the optimization process with a low weight $\lambda_{1,2}=0.001$. We ran ROS with 150 epochs for Stage I and 45 for Stage II. Each experiment is repeated three times taking the result on the target at the last epoch.

\noindent It is worth noting that we essentially use the same set of parameters for all settings. This highlights that our method can generalize across datasets and network architectures without specific finetuning of the hyper-parameters. For the sake of completeness, we also provide a fully detailed evaluation of ROS including the OS metric and standard deviation for all our experiments in Tables \ref{tab:officehomeResnet50sup} and \ref{tab:office31Resnet50sup}. We remark that, in terms of OS and OS*, STA is extremely unstable with large standard deviations over multiple runs.

\section{Reproducibility Study}
We extend here the reproducibility study presented in the main paper considering also further results on the Office-Home dataset. Specifically, in Table \ref{tab:reproducibilitysup} we compare the results published in the official papers of STA \cite{liu2019separate},  OSBP \cite{saito2018open}, and UAN \cite{you2019universal} considering the OS accuracy since it is the only metric shared by all the works. For UAN we replicated the particular settings described in the original publication: for Office-Home the first 10 classes in alphabetic order are shared between source and target, the next five are private source classes and all the others are private target classes. For Office-31 the first 10 classes in alphabetic order are shared between source and target, the next 10 are private source classes and all the others are private target classes. It is worth noting that, although we used the code provided by the authors and we followed the instructions provided in the related papers, the obtained results are lower than the declared ones, with gaps that range between $1.9\%$ and $6.2\%$.

\begin{table}[t]
\centering
\caption{Reported vs reproduced OS accuracy (\%) averaged over three runs on all the sub-domains of  Office-31 and Office-Home with the indicated backbones.}
\resizebox{\textwidth}{!}{
\begin{tabular}{l@{~~~} cccccc ccc cccccc}
\hline
 \multicolumn{15}{c}{\textbf{Reproducibility Study}} \\
 \hline
 \multicolumn{6}{c|}{Office-31 (ResNet-50)} & \multicolumn{3}{c|}{Office-31 (VGGNet)} & \multicolumn{6}{c}{Office-Home  (ResNet-50)}\\
 \hline
  \multicolumn{3}{c|}{STA\textsubscript{sum}} & \multicolumn{3}{c|}{UAN}  & \multicolumn{3}{c|}{OSBP} & \multicolumn{3}{c|}{STA\textsubscript{sum}} & \multicolumn{3}{c}{UAN}\\

OS\textsubscript{reported} & OS\textsubscript{ours} & \multicolumn{1}{c|}{gap} & OS\textsubscript{reported} & OS\textsubscript{ours}  & \multicolumn{1}{c|}{gap} & OS\textsubscript{reported} & OS\textsubscript{ours}  & \multicolumn{1}{c|}{gap} & OS\textsubscript{reported} & OS\textsubscript{ours}  & \multicolumn{1}{c|}{gap} & OS\textsubscript{reported} & OS\textsubscript{ours}  & \multicolumn{1}{c}{gap}\\
 \hline

\multicolumn{1}{c}{92.9} &  \multicolumn{1}{c}{90.6$\pm$1.8} & \multicolumn{1}{c|}{\textbf{2.3}} &  \multicolumn{1}{c}{89.2} &   \multicolumn{1}{c}{87.9$\pm$0.03} & \multicolumn{1}{c|}{\textbf{1.3}} &  \multicolumn{1}{c}{89.1} &  \multicolumn{1}{c}{84.2 $\pm$0.4} & \multicolumn{1}{c|}{\textbf{4.9}} & \multicolumn{1}{c}{69.5} &   \multicolumn{1}{c}{63.3$\pm$2.1} & \multicolumn{1}{c|}{\textbf{6.2}}  &  \multicolumn{1}{c}{77.0} & \multicolumn{1}{c}{75.1 $\pm$0.2} & \multicolumn{1}{c}{\textbf{1.9}}\\

\hline
\end{tabular}
}
\hspace{-2cm}
\label{tab:reproducibilitysup} 
\end{table}

For complete transparency, we summarize here all the details about implementation, code and hyper-parameters used for running the competitor methods. 

\paragraph{STA} \cite{liu2019separate}          \url{https://github.com/thuml/Separate_to_Adapt}\\
The code provides a full description of how to run STA for the the A$\rightarrow$D domain shift of Office-31 with ResNet-50 backbone. For the experiments on Office-31 we trained for $900$ iterations in Stage I ($400$ for the multi-binary classifier and $500$ for the known/unknown classifier) and $1900$ iterations in Stage II. We used batch size $32$, SGD with momentum $0.9$ and weight decay $0.0005$. We used the inverse scheduling for the learning rate that is set as  $0.001$ in Stage I and $0.0005$ in Stage II ($10$ times smaller for finetuned layers). Since the paper does not differentiate between Office-31 and Office-Home in terms of hyper-parameters, we ran the experiments on Office-Home with the same exact values.

It is worth noting that there is some ambiguity around the value of the learning rate for STA. The paper indicates that the learning rate may be adjusted in the $\{0.001,1\}$  range with cross-validation. However the code does not provide any validation routine, thus it is unclear how this parameter should be further refined. In addition, the learning rate set for Stage II in the code is outside the range indicated in the paper. 
In our experiments, we kept always the learning rate set in the code. We also found other ambiguities between the paper and the code. The paper indicates that in Stage I the feature extractor is trained on the source samples, while the feature extractor is frozen with the original weights from ImageNet. Moreover, as already discussed in our main submission, the paper presents a similarity score based on the \emph{max} operator, while it is implemented with a \emph{sum} operator in the released code. Finally, although the paper includes results with VGGNet, the code for this variant is not provided, nor specific details are discussed in the paper, which prevents reproducibility.

\paragraph{OSBP} \cite{saito2018open} \url{https://github.com/ksaito-ut/OPDA_BP}\\
This repository provides the code for OSBP, both with the VGGNet and ResNet-50 backbones.
Specifically, the instructions explain how to run OSBP on the VisDA-2017 dataset \cite{visda2017} with VGG-19. 
For the experiments using VGGNet on Office-31, we used the provided implementation and we followed the description of the OSDA paper in using batch size $32$, SGD with momentum $0.9$, learning rate $0.001$, and weight decay $0.0005$. We trained only the new layers for $500$ epochs, while the others were frozen with ImageNet weights. For the experiments using ResNet-50, we use batch size $32$, learning rate $0.001$, and train for $300$ epochs for Office-Home and $500$ for Office-31.
Since the authors mentioned that the library version can make a significant difference in the results,
for all the experiments we used exactly their declared version (Pytorch 0.3).

\paragraph{UAN} \cite{you2019universal} \url{https://github.com/thuml/Universal-Domain-Adaptation}\\
This repository provides the code for UAN with ResNet-50 as backbone:  specific files contain instructions to run experiments on both Office-31 and Office-Home. On Office-31 we trained for $20000$ iterations with batch size $36$, SGD with and momentum $0.9$, learning rate $0.001$ for new layers and $0.0001$ for finetuned layers with inverse scheduling, and weight decay $0.0005$. On Office-Home we trained for $40000$ iterations with batch size $36$, SGD with and momentum $0.9$, learning rate $0.01$ for new layers and $0.001$ for finetuned layers with inverse scheduling, and weight decay $0.0005$. It is worth noting that the original evaluation implemented in the code would have saved the performance of UAN on the test data at each epoch and presented the best accuracy (OS) at the end of the training. This is not a standard procedure. To avoid its possibly unfair beneficial effect we provide the results obtained after the last epoch as done for all the other benchmark methods in our experiments.

\section{Extended Openness Analysis}
\begin{figure}[t!]
\begin{tabular}{ccc}
\includegraphics[width=0.33\linewidth]{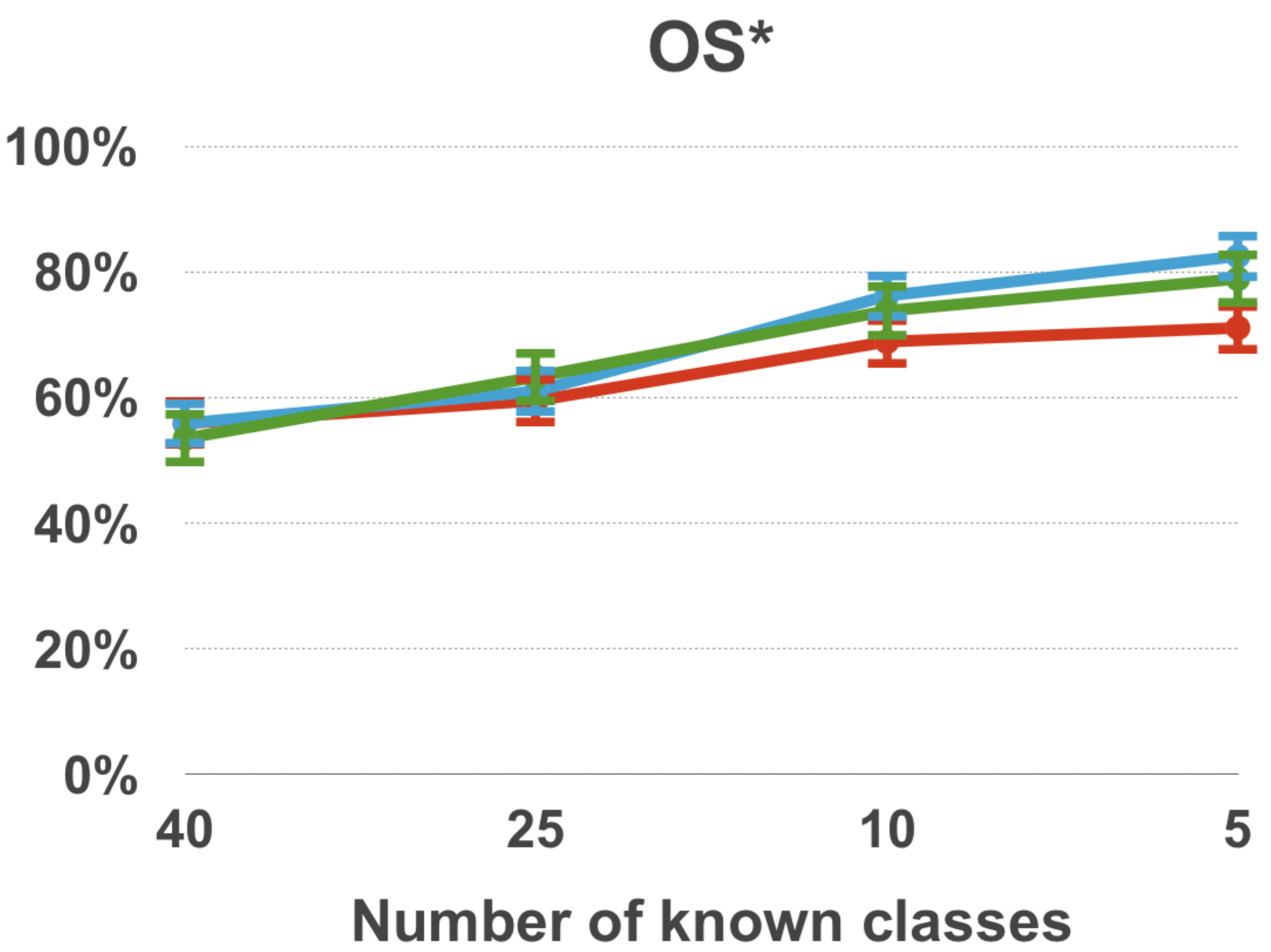} &
\includegraphics[width=0.33\linewidth]{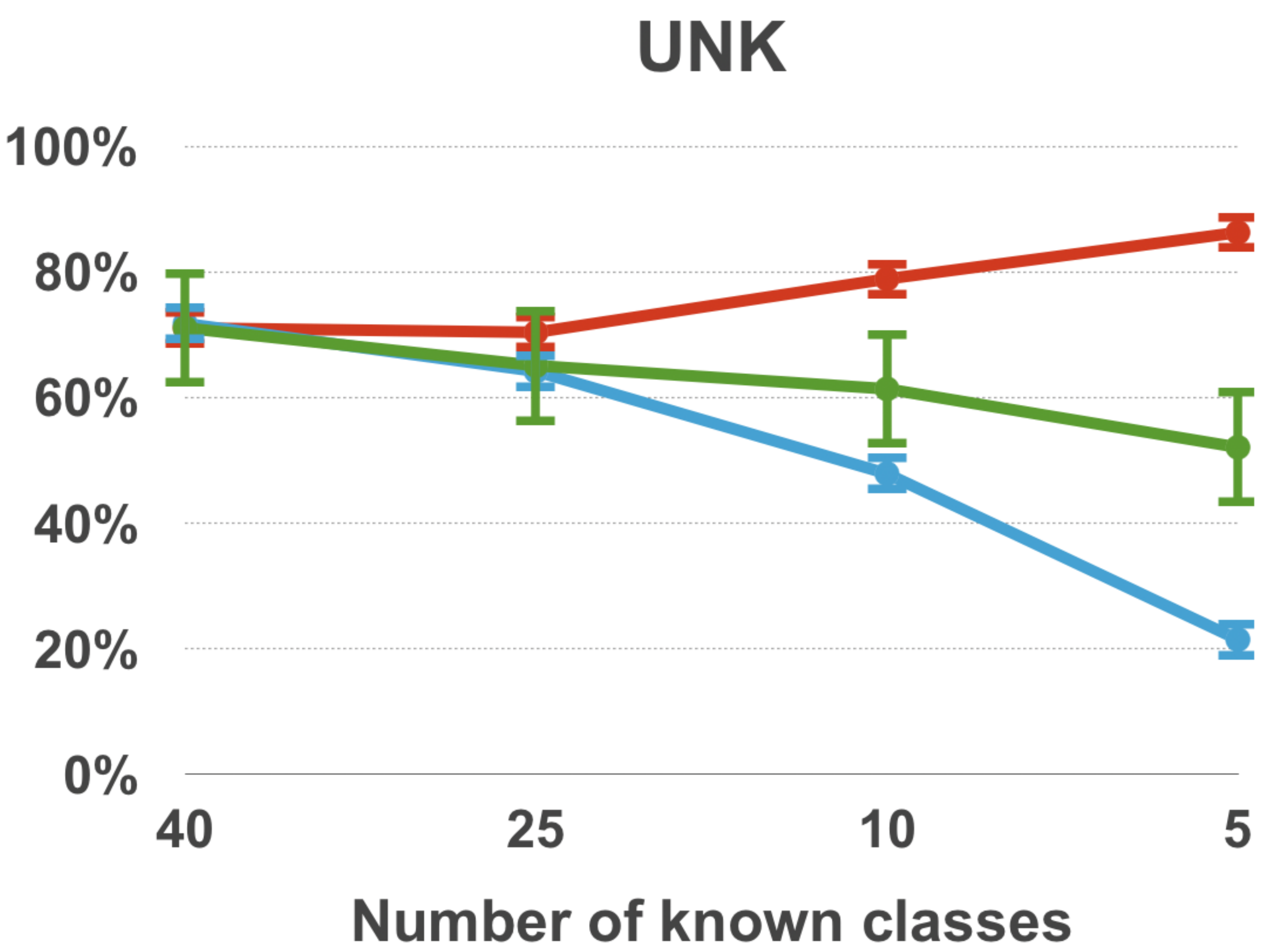} &
\includegraphics[width=0.33\linewidth]{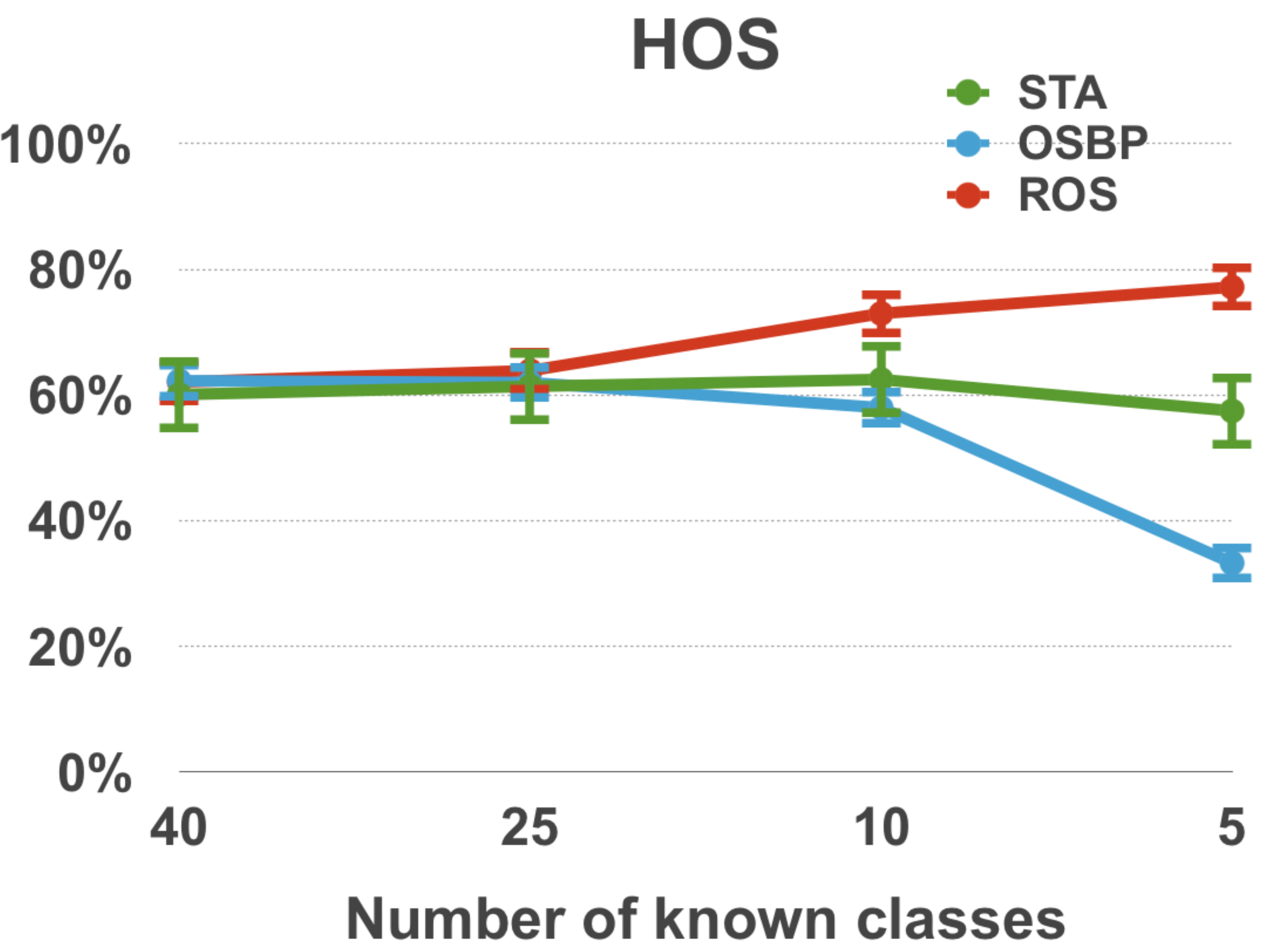} \\
\end{tabular}
\resizebox{\textwidth}{!}{
\begin{tabular}{l@{~~~~}ccccc cccc}
\hline
       \multicolumn{9}{c}{\textbf{Office-Home (Avg.)}}        \\
       \hline
       & \multicolumn{4}{c|}{40 known classes} & \multicolumn{4}{c}{25 known classes}\\ \hline
       & OS & OS* & UNK & \multicolumn{1}{c|}{\textbf{\underline{HOS}}} & OS & OS* & UNK & \multicolumn{1}{c}{\textbf{\underline{HOS}}}\\
\hline

STA\textsubscript{sum}\cite{liu2019separate} & 53.9$\pm$1.7 & 53.5$\pm$1.7 & 71.1$\pm$3.0 & \multicolumn{1}{c|}{60.1$\pm$1.7} & 63.5$\pm$2.6 & 63.3$\pm$3.3 & 65.1$\pm$2.2 & \multicolumn{1}{c}{61.4$\pm$1.5}\\
OSBP\cite{saito2018open} & 56.3$\pm$1.6 & 55.9$\pm$1.6  & 71.8$\pm$3.6 & \multicolumn{1}{c|}{\textbf{62.3$\pm$1.8}} & 61.4$\pm$2.5 & 61.0$\pm$2.8 & 64.2$\pm$2.8 & \multicolumn{1}{c}{62.0$\pm$2.8}\\
\textbf{ROS} & 56.2$\pm$1.2 & 55.9$\pm$1.2 & 71.1$\pm$2.8 & \multicolumn{1}{c|}{62.1$\pm$1.8}  & 59.9$\pm$1.9 & 59.5$\pm$1.9 & 70.4$\pm$4.0 & \multicolumn{1}{c}{\textbf{63.9$\pm$2.7}} \\
\hline
& \multicolumn{4}{c|}{10 known classes} & \multicolumn{4}{c}{5 known classes}\\ \hline
 & OS & OS* & UNK & \multicolumn{1}{c|}{\textbf{\underline{HOS}}} & OS & OS* & UNK & \multicolumn{1}{c}{\textbf{\underline{HOS}}}\\
 \hline
 STA\textsubscript{sum}\cite{liu2019separate} & 72.5$\pm$5.8 & 73.8$\pm$6.6 & 61.4$\pm$14.5 & \multicolumn{1}{c|}{62.5$\pm$8.6} & 74.4$\pm$2.8 & 78.9$\pm$6.3 & 52.1$\pm$17.7 & \multicolumn{1}{c}{57.5$\pm$12.3}\\
 OSBP\cite{saito2018open} & 73.6$\pm$5.7 & 76.2$\pm$5.7 & 47.9$\pm$4.8 & \multicolumn{1}{c|}{58.0$\pm$5.1} & 72.3$\pm$4.9 & 82.5$\pm$5.6 & 21.5$\pm$1.6 & \multicolumn{1}{c}{33.3$\pm$2.6}\\
 \textbf{ROS} & 69.9$\pm$5.2 & 68.9$\pm$5.5 & 78.9$\pm$2.9 & \multicolumn{1}{c|}{\textbf{73.0$\pm$4.4}} & 73.4$\pm$6.7 & 71.1$\pm$7.6 & 86.3$\pm$2.7 & \multicolumn{1}{c}{\textbf{77.2$\pm$5.9}} \\
 
\hline
\end{tabular}
}
\caption{Accuracy (\%) averaged over the three configurations designed for each degree of openness considered: with 40, 25, 10 and 5 known classes. The table reports in details the values used to prepare the plots}
\label{fig:decreaseknownclasssup}
\end{figure}

Following the openness analysis of Figure 4 of the main paper, we also extend the evaluation to include a case with lower openness: \textbf{40} known classes $\bbO=1-\frac{40}{65}=0.38$ using ID:\{0-39, 15-54, 25-64\}. The results in Figure \ref{fig:decreaseknownclasssup} confirms the trend already observed in the main paper. Given the low UNK and HOS results of UAN we did not include this method in the ablation and focused only on the two best competitors of ROS: OSBP and STA.

\section{Sensitivity analysis of the hyper-parameters}
\begin{figure}[t!]
\begin{tabular}{cccc}
\includegraphics[width=0.25\linewidth]{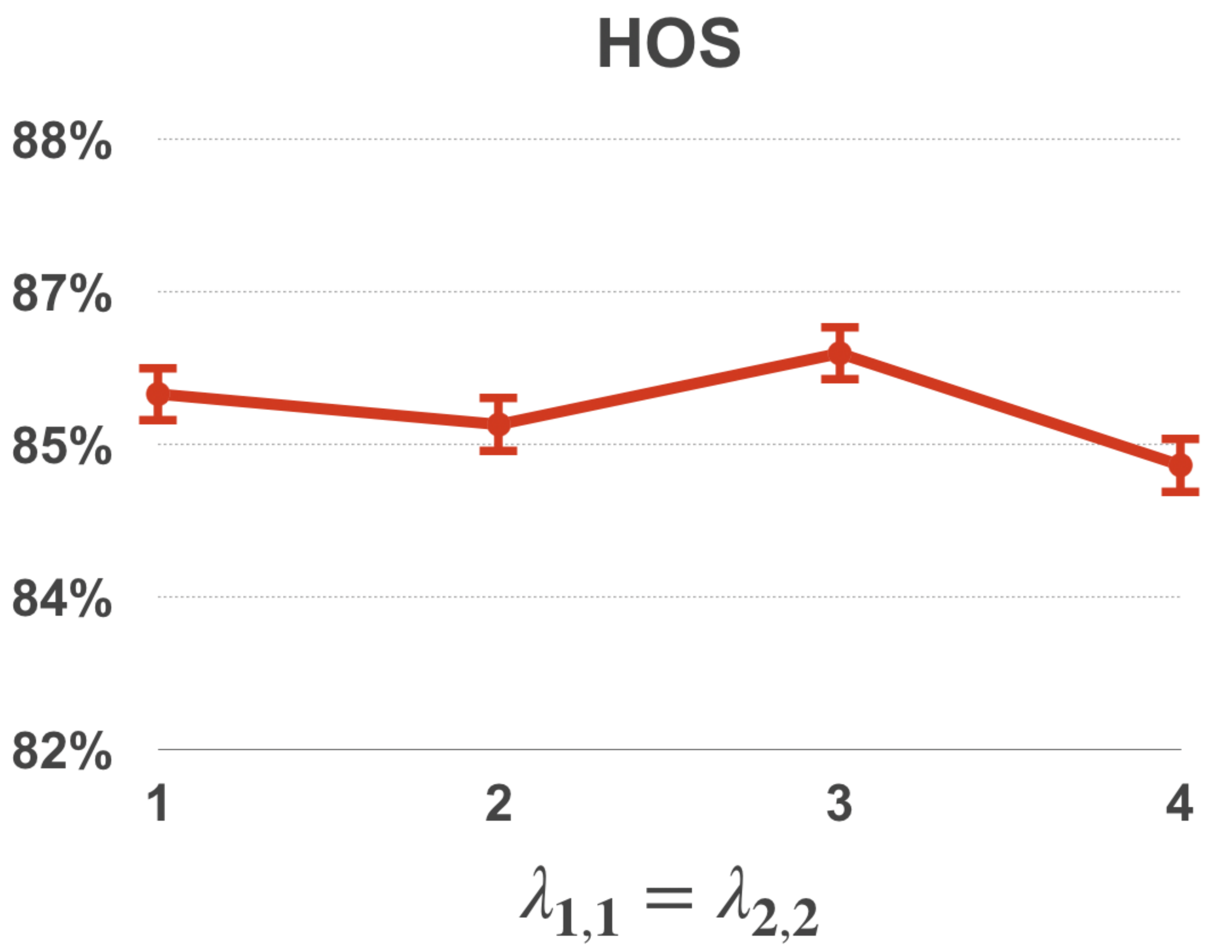} &
\includegraphics[width=0.25\linewidth]{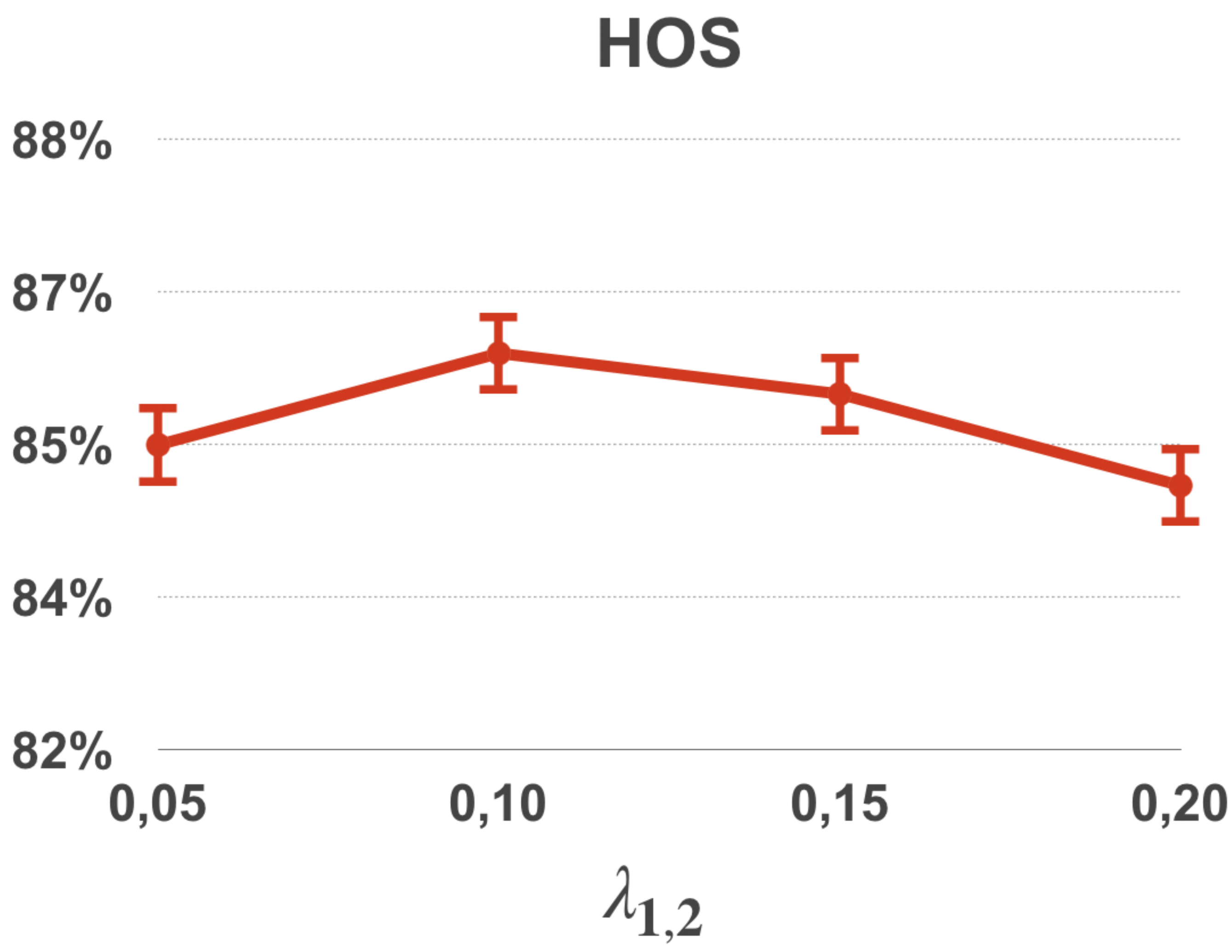} &
\includegraphics[width=0.25\linewidth]{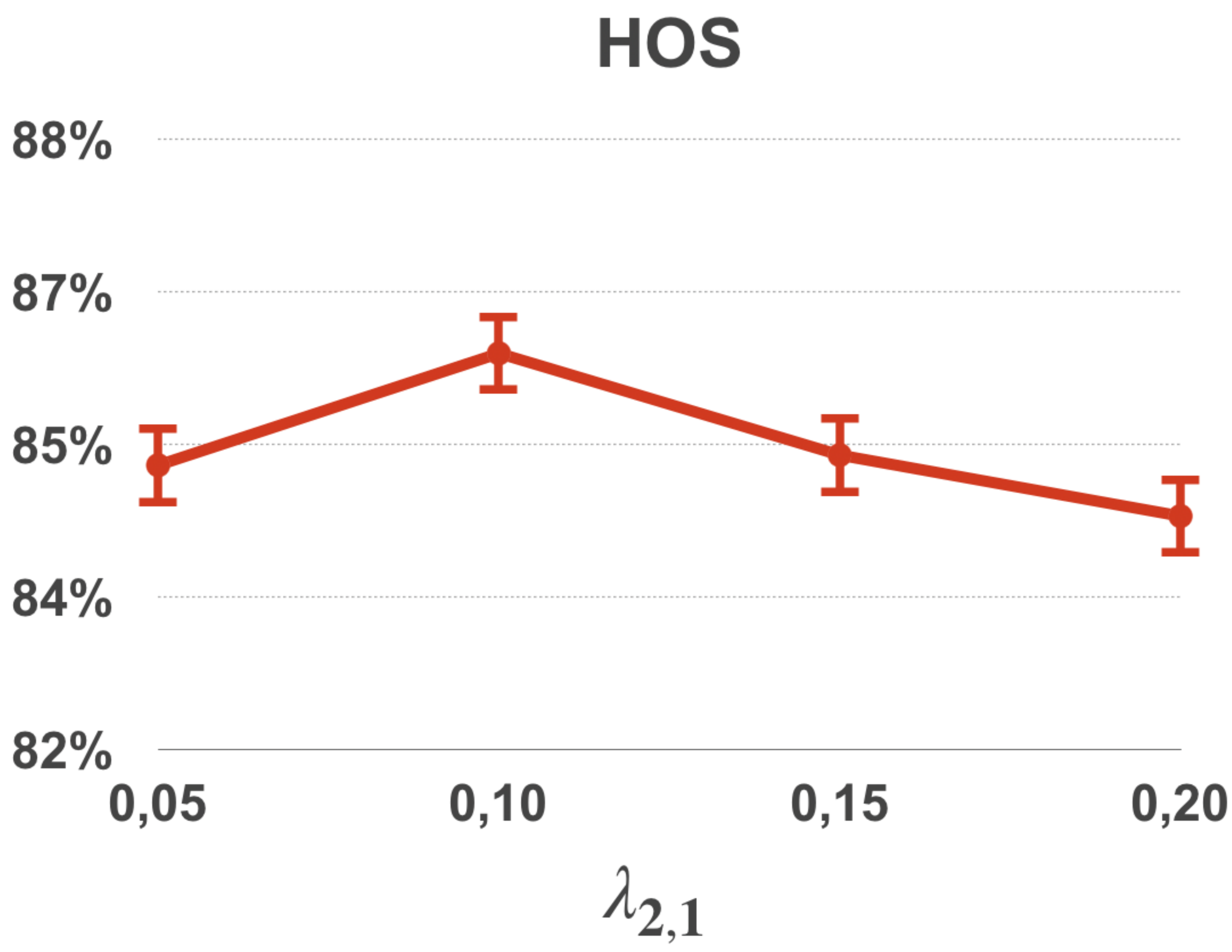} &
\includegraphics[width=0.25\linewidth]{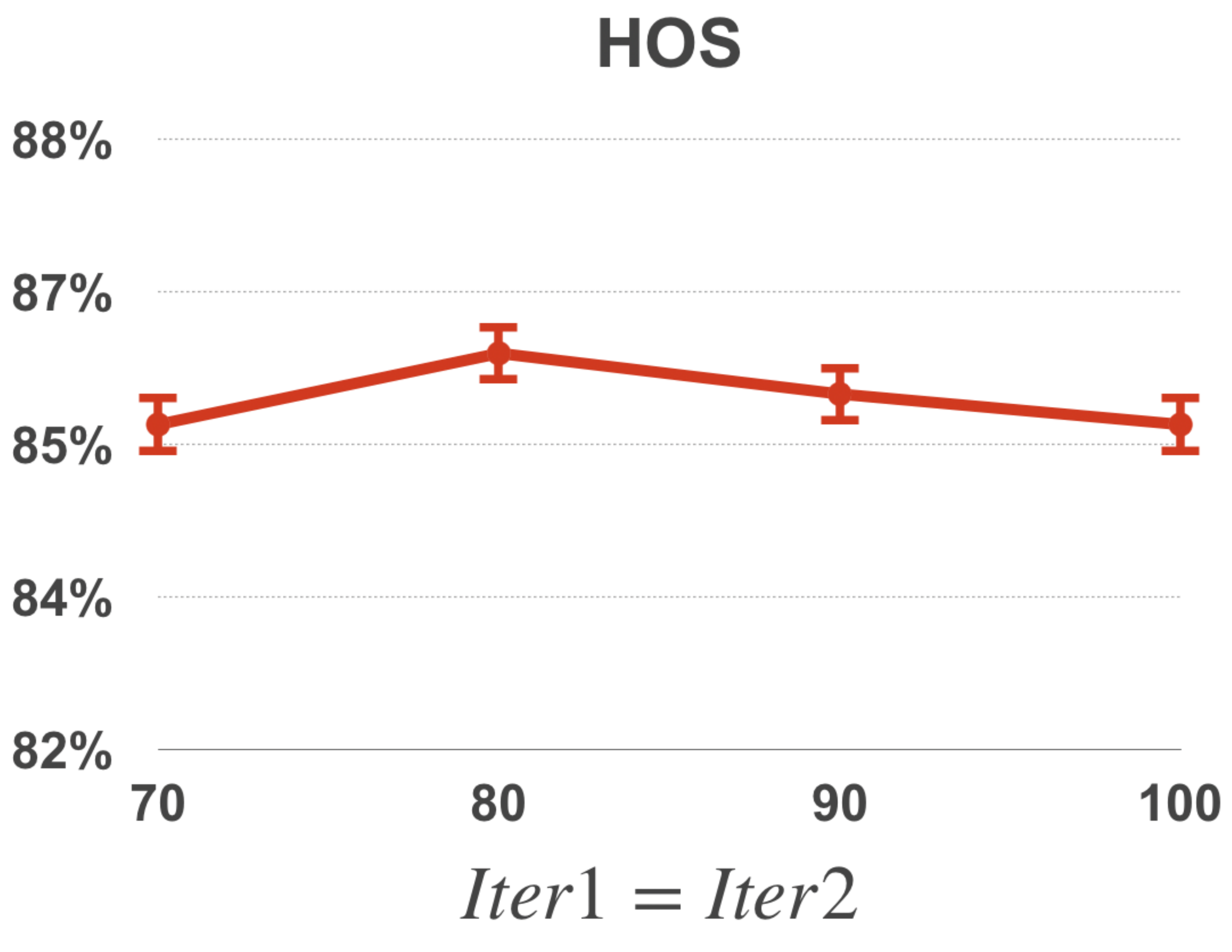}\\
\end{tabular}
\caption{Hyper-parameter analysis}
\label{fig:hp_analysis}
\end{figure}

We perform a sensitivity analysis to evaluate the impact of changes in the hyper-parameter values on the performance of ROS. The experiments are performed on Office-31 with ResNet-50 as backbone and the results are displayed in Figure \ref{fig:hp_analysis}. ROS is not very sensitive to the value of the hyper-parameters, with only $\lambda_{2,1}$ causing a variation in HOS $>1.0$. Please note that it is safe to set the entropy weight to 0.1 without hyper-parameters tuning, exactly as done in \cite{Carlucci_2019_CVPR,liu2019separate,Xu_2019_ICCV}. Regardless of the specific hyper-parameters, ROS outperforms its best competitor OSBP (HOS=83.7) confirming that the superior performance is the result of algorithmic novelty rather than from hyper-parameters tuning. Moreover, we underline that we use the same hyper-parameters for all 18 domain pairs demonstrating that the choice of the hyper-parameters' value is robust across datasets.
As a final remark, we note that ROS has a comparable number of parameters with respect to competing approaches. Indeed $\lambda_{1,1}$ and $\lambda_{2,2}$ are defined separaterly, but they are in fact constrained to the same value. So overall ROS has three parameters and two for the training iterations, the same as the most recent AoD (see Equation (3) of \cite{feng2019attract}).

\begin{table}[t!]
\centering
\caption{Analysis on the use of self-supervised tasks for the two stages of the method and further ablation.}
\resizebox{\textwidth}{!}{
\begin{tabular}{l@{~} c c c c c c c c}

\hline
\multicolumn{8}{c}{\textbf{Other Self-Supervised Tasks \& Ablation Study}}\\
\hline
\multicolumn{1}{c}{\multirow{1}{*}{\textbf{STAGE I} (AUC-ROC)}}  & \multicolumn{1}{c|}{A $\rightarrow$ W } & \multicolumn{1}{c|}{A $\rightarrow$ D } & \multicolumn{1}{c|}{D $\rightarrow$ W }  & \multicolumn{1}{c|}{W $\rightarrow$ D} & \multicolumn{1}{c|}{D $\rightarrow$ A } & \multicolumn{1}{c|}{W $\rightarrow$ A } & \multicolumn{1}{c}{\textbf{Avg.}} \\
\hline

\textbf{ROS} & \multicolumn{1}{c|}{90.1} & \multicolumn{1}{c|}{88.1} & \multicolumn{1}{c|}{99.4} & \multicolumn{1}{c|}{99.9} & \multicolumn{1}{c|}{87.5} & \multicolumn{1}{c|}{83.8} & \textbf{91.5}\\
ROS - Translation & \multicolumn{1}{c|}{80.8} & \multicolumn{1}{c|}{74.9} & \multicolumn{1}{c|}{82.2} & \multicolumn{1}{c|}{98.8} & \multicolumn{1}{c|}{72.0} & \multicolumn{1}{c|}{79.1} & 81.3 \\

ROS - Rotation+Translation & \multicolumn{1}{c|}{82.4} & \multicolumn{1}{c|}{79.3} & \multicolumn{1}{c|}{99.0} & \multicolumn{1}{c|}{99.4} & \multicolumn{1}{c|}{82.6} & \multicolumn{1}{c|}{82.8} & 87.6 \\

ROS - 4-Class Rotation & \multicolumn{1}{c|}{58.7} & \multicolumn{1}{c|}{57.2} & \multicolumn{1}{c|}{70.0} & \multicolumn{1}{c|}{78.4} & \multicolumn{1}{c|}{55.8} & \multicolumn{1}{c|}{56.9} & 62.9 \\

 \hline\hline
\multicolumn{1}{c}{\multirow{1}{*}{\textbf{STAGE II} (HOS)}}  & \multicolumn{1}{c|}{A $\rightarrow$ W } & \multicolumn{1}{c|}{A $\rightarrow$ D } & \multicolumn{1}{c|}{D $\rightarrow$ W }  & \multicolumn{1}{c|}{W $\rightarrow$ D} & \multicolumn{1}{c|}{D $\rightarrow$ A } & \multicolumn{1}{c|}{W $\rightarrow$ A } & \multicolumn{1}{c}{\textbf{Avg.}} 
 \\
\hline
\textbf{ROS}  & \multicolumn{1}{c|}{82.1}  & \multicolumn{1}{c|}{82.4}  & \multicolumn{1}{c|}{96.0}  & \multicolumn{1}{c|}{99.7}   & \multicolumn{1}{c|}{77.9}  & \multicolumn{1}{c|}{77.2}  & \textbf{85.9}   \\
ROS - Jigsaw  & \multicolumn{1}{c|}{83.1}  & \multicolumn{1}{c|}{79.3}  & \multicolumn{1}{c|}{93.5}  & \multicolumn{1}{c|}{100.0}   & \multicolumn{1}{c|}{75.5}  & \multicolumn{1}{c|}{76.1}  & 84.6   \\

ROS - Rotation+Jigsaw  & \multicolumn{1}{c|}{85.7}  & \multicolumn{1}{c|}{80.5}  & \multicolumn{1}{c|}{95.0}  & \multicolumn{1}{c|}{100.0}   & \multicolumn{1}{c|}{76.0}  & \multicolumn{1}{c|}{76.7}  & 85.7   \\

ROS Stage I - $\lambda_{2,1}=0$ Stage II  & \multicolumn{1}{c|}{79.4}  & \multicolumn{1}{c|}{82.0}   & \multicolumn{1}{c|}{95.3}  & \multicolumn{1}{c|}{99.6} & \multicolumn{1}{c|}{75.1} & \multicolumn{1}{c|}{72.5} &  84.0\\

ROS Stage I - ROS Stage II+Center Loss & \multicolumn{1}{c|}{79.6}  & \multicolumn{1}{c|}{82.8}   & \multicolumn{1}{c|}{95.1}  & \multicolumn{1}{c|}{99.5} & \multicolumn{1}{c|}{77.8} & \multicolumn{1}{c|}{76.3} &  85.2\\

\hline
\end{tabular}
}
    \label{tab:ablationstep1sup} 
\end{table}

\begin{table}[t!]
\centering
\caption{Runtime analysis on Office-31(A-W) with ResNet-50. Hardware - CPU: Intel(R) Core(TM) i7-5930K @ 3.50GHz, GPU (x1): Nvidia GeForce GTX 1080Ti.}
\begin{tabular}{l@{~} c @{~~~~} c @{~~~~} c @{~~~~} c  }
\hline
\multicolumn{4}{c}{\textbf{Time analysis}}\\
\hline
\multicolumn{1}{c@{~~~~}}{STA\cite{liu2019separate}}  & \multicolumn{1}{c@{~~~~}}{UAN\cite{you2019universal}} & \multicolumn{1}{c@{~~~~}}{OSBP\cite{saito2018open}} & \multicolumn{1}{c@{~~~~}}{\textbf{ROS}}
 \\
\hline
1069s & 9615s & 3672s & 1875s \\
\hline
\end{tabular}
    \label{tab:time} 
\end{table}

\begin{algorithm}[b!]
\caption{Compute normality score and Generate $\sD_t^{knw}$ \& $\sD_t^{unk}$}
\label{alg:score}
\begin{algorithmic}
\Require 
\State Trained networks $E$ and $R_1$
\State Target dataset $\sD_t=\{\bx^t_j\}_{j=1}^{N_t}$
\Ensure
\State Known target dataset $\sD_t^{knw}=\{\bx_j^{t,knw}\}_{j=1}^{N_{t,knw}}$
\State Unknown target dataset $\sD_t^{unk}=\{\bx_j^{t,unk}\}_{j=1}^{N_{t,unk}}$
\State 
\Procedure{getRotationScore}{$\bz$,$i$}
\State $\textbf{o}=zeros(|\sC_s|)$ \# vector of $|\sC_s|$ zeros
\For{\textbf{each} $k$ \textbf{in} $\{1,...,|\sC_s|\}$}
\State $[\textbf{o}]_{k}=[\bz]_{k\times4+i}$ \# $[\textbf{a}]_b$ indicated the $b$-th element of vector $\textbf{a}$
\EndFor
\State \textbf{return} $\textbf{o}$
\EndProcedure
\Procedure{getEntropyScore}{$\bz$}
\State \textbf{return} $\bz \cdot \log(\bz)/\log(|\sC_s|)$
\EndProcedure
\Procedure{getNormalityScore}{$E$,$R_1$,$\sD_t$}
\For{\textbf{each} $\bx^t_j$ \textbf{in} $\sD_t$}
\State Initialize: $h=\{\}$,  $\textbf{o}=zeros(|\sC_s|)$ 
\For{\textbf{each} $i$ \textbf{in} $\{1,...,4\}$}
\State $\tilde{\bx}_j = rot90(\bx_j,i)$
\State $\bz_j=\text{softmax}\big(R_1(E(\bx_j)||E(\tilde{\bx}_j))\big)$
\State $h \leftarrow getEntropyScore(\bz_j)$
\State $o\mathrel{+}=getRotationScore(\bz_j,i)$ \# element-wise sum of vectors
\EndFor
\State $h = mean(h)$
\State $o = max(\textbf{o})$
\State $\mathcal{N} \leftarrow \eta_j = max(o,1-h)$
\EndFor
\State \textbf{return} $\mathcal{N}$
\EndProcedure
\State
\Procedure{Main}{ }
\State Initialize: $\sD_t^{knw}=\{\}$, $\sD_t^{unk}=\{\}$
\State $\sA = getNormalityScore(E,R_1,\sD_t)$
\For{\textbf{each} $(\bx_j,\eta_j)$ \textbf{in} $(\sD_t,\mathcal{N})$}
\If{$\eta_j \ge mean(\mathcal{N})$}
\State $\sD_t^{knw} \leftarrow \bx_j$
\Else
\State $\sD_t^{unk} \leftarrow \bx_j$
\EndIf
\EndFor
\EndProcedure
\end{algorithmic}
\end{algorithm}

\section{Other Self-Supervised Tasks and Further Ablation}
Our goal is to show that it is possible to successfully tackle both sub-tasks of OSDA, known/unknown separation
and domain alignment, with a single self-supervised task. From the literature of CSDA \cite{xu2019self-supervised,Carlucci_2019_CVPR} and
anomaly detection \cite{golan2018deep,Bergman2020Classification-Based,hendrycks2019selfsupervised}, rotation classification clearly emerges as the most reliable candidate for our purpose.
To confirm our claim, we run additional experiments on Office-31 (ResNet-50) with alternative self-supervised tasks.
Following \cite{golan2018deep}, we considered the self-supervised task of translation classification for anomaly detection (Stage I). Moreover, following \cite{Carlucci_2019_CVPR}, we considered the self-supervised task of solving a jigsaw puzzle for domain alignment (Stage II).
Table \ref{tab:ablationstep1sup} show the obtained results: in both sets of experiments, rotation recognition alone outperforms both the alternative task and combination of the two tasks.

We also confirm the crucial contribution of the multi-rotation task instead of the standard 4-Class task in Stage I. Table \ref{tab:ablationstep1sup} shows that the standard rotation decreases the AUC-ROC by an astonishing $28.6\%$. Of course we keep the anchor (relative rotation) also in this 4-Class experiment.

Since using the entropy loss in the object classification process across domains is standard practice, we did not include an ablation for Stage II of ROS on this term in the main paper. For completeness we present it here. We set $\lambda_{2,1}=0$ including the results in Table \ref{tab:ablationstep1sup}: as expected, without the entropy loss the performance drop on average of 1.9 percentage points, confirming that the entropy helps to adapt with a more evident effect in case of large domain gaps (\eg A$\rightarrow$W, W$\rightarrow$A).
Moreover, in Stage II, the center loss is not as relevant as for Stage I, and it would imply the introduction of an extra hyper-parameter. Indeed, the results in in Table \ref{tab:ablationstep1sup} indicate that adding the center loss to Stage II might even produce a slight drop in performance.

\section{Time analysis}
We executed a training runtime analysis on Office-31(A-W) with ResNet-50 for all the methods discussed in the paper with their indicated hyper-parameters. The results in Table \ref{tab:time} show that that the time is not an issue and ROS is even twice as fast as its best competitor in terms of HOS performance (OSBP).

\section{Normality Score Pseudo-code}
As promised in the main paper we summarize in Algorithm \ref{alg:score} the procedure used to calculate the normality score at the end of Stage I of ROS.

\bibliographystyle{splncs04}
\bibliography{egbib.bib}

\end{document}